\documentclass{article}

\usepackage[preprint]{neurips_2026}

\usepackage[utf8]{inputenc} 
\usepackage[T1]{fontenc}    
\usepackage{hyperref}       
\usepackage{url}            
\usepackage{enumitem}       
\usepackage{booktabs}       
\usepackage{amsfonts}       
\usepackage{nicefrac}       
\usepackage{microtype}      
\usepackage{xcolor}         
\usepackage{graphicx}       
\usepackage{xurl}
\usepackage{caption}
\title{Cyclic Denoising Reveals Ultrastable Memories in Diffusion Models}

\author{%
  Rishabh Sharma$^{1, 2}$\thanks{Corresponding author} \qquad
  Stefano Martiniani$^{1, 2, 3, 4}$ \\[0.8em]
  $^{1}$Simons Center for Computational Physical Chemistry, New York University \\
  $^{2}$Center for Soft Matter Research, Department of Physics, New York University \\
  $^{3}$Center for Neural Science, New York University \\
  $^{4}$Courant Institute of Mathematical Sciences, New York University \\[0.5em]
  \texttt{\{rs10125, sm7683\}@nyu.edu}
}
\begin{document}

\maketitle

\begin{abstract}
We introduce cyclic denoising—repeated forward and reverse diffusion at controlled noise amplitudes—as an extraction attack for image diffusion models. Inspired by random organization in disordered solids, where cyclic mechanical perturbations anneal the system into increasingly stable configurations, cyclic denoising exposes regions of the learned distribution that remain largely inaccessible to standard sampling. We find that these dynamics drive samples toward attractors with a broad stability spectrum, with the deepest attractors exhibiting ultrastability: they can be regenerated from near-total corruption and sustained through thousands of noising--denoising cycles. Many of these deep attractors correspond to memorized training images, including stock photographs, brand watermarks, and web-crawl artifacts. Our extraction attack requires only sampler-level control—the ability to partially noise a sample to an intermediate diffusion timestep and denoise it back—but no gradients and no weight inspection. Crucially, cyclic denoising requires no prior knowledge of training data, captions, or prompts. In contrast, prior generate-and-filter attacks on production-scale diffusion models commonly rely on large-scale prompted generation from known or suspected training captions, followed by post-hoc similarity search or membership-inference filtering to identify memorized candidates. While cyclic denoising can also be applied with prompts, our main protocol is fully unconditioned. We demonstrate the phenomenon in Stable Diffusion v1.4, a latent diffusion model, and in a smaller pixel-space DDPM, showing consistent behavior across latent- and pixel-space diffusion models. Across noise amplitudes, we observe a yielding-like transition: low-amplitude cycling produces either trivial absorbing fixed points (featureless, near-monochromatic images) or limit cycles (traveling/oscillating Turing-like patterns in image space), while larger amplitudes induce rearrangements, basin hopping, and long-lived trapping in structured memorized attractor basins. We further observe hierarchical partial absorption, where coarse scene layout freezes while fine details remain diffusive, as well as prompt-stabilized basins and cross-initial-condition universality of the recovered attractor set. Together, these results establish cyclic denoising as both a physics-inspired probe of generative landscapes and a practical tool for memorization auditing, with implications for privacy, copyright compliance, and model fingerprinting.

\end{abstract}

\section{Introduction}
Large generative models are trained on massive collections of web-scale data. Their capabilities depend on these datasets, but so do their risks: models may retain and reproduce specific training examples, including copyrighted artwork, proprietary media, personal photographs, watermarked stock images, and other sensitive artifacts scraped from the web. Understanding when and how such data can be recovered from a trained model is therefore central to privacy, copyright compliance, and model auditing.

Existing extraction attacks on diffusion models have shown that memorized images can sometimes be recovered, but they rely in different ways on signals or machinery external to the model's sampling dynamics. One canonical, computationally expensive approach is generate-and-filter: for text-conditioned models such as Stable Diffusion, attacks prompt the model with captions known or suspected to correspond to training examples, generate many samples per caption, and apply post-hoc filtering, clustering, similarity search, or membership-inference procedures to isolate memorized outputs~\citep{carlini2023extractingtrainingdatadiffusion}. For unconditional models, where prompts are unavailable, memorized candidates have instead been identified by comparing large sets of generated samples to the training set, a strategy mainly practical in benchmark-scale settings where the training set is known and searchable~\citep{carlini2023extractingtrainingdatadiffusion}. Subsequent attacks use different signals, such as one-step denoising behavior~\citep{webster2023reproducibleextractiontrainingimages} or auxiliary classifiers~\citep{chen2025sidesurrogateconditionaldata}, but still rely on captions, training-data access, or separately trained detectors. Cyclic denoising instead uses the sampler's own long-time dynamics to propose memorized candidates: absorbing episodes and long residence times, not captions, training-set lookup, or auxiliary detectors, determine what is inspected.

Our approach is inspired by the physics of driven disordered systems~\citep{Corte2008,Sharma2025,mungan2025,zhang2026absorbing}. 
The setup is conceptually simple: a disordered solid, for example, has a rugged landscape with many metastable configurations and can be perturbed cyclically at a controlled amplitude, such as by repeated shear. 
Configurations change from cycle to cycle until the dynamics reach a state that resists further change, an absorbing state. 
The amplitude of the drive controls which absorbing states are reachable. 
Weak perturbations leave the system trapped in shallow basins; stronger perturbations dislodge it from those basins and let it explore deeper, more stable regions of the landscape. At even higher amplitudes, the system yields and ergodically explores the energy landscape. Thus, an absorbing to diffusive transition is observed with increasing cyclic perturbation amplitude.  

We import this idea into diffusion models through \emph{cyclic denoising}. Starting from an image or latent, we repeatedly apply forward noising to a controlled amplitude $\gamma$ and then reverse denoise back to $\gamma=0$, using the output of one cycle to initialize the next. Viewed stroboscopically, i.e., at the end of each completed noising--denoising cycle, the model traces a trajectory through its learned generative landscape. The diffusion noise level $\gamma$ plays the role of a drive amplitude: by varying $\gamma$, we tune the strength of the cyclic perturbation and probe which states are reached and remain stable. Long plateaus of near-unity stroboscopic cosine similarity mark absorbing episodes; decoding these states yields candidate memorized images.\footnote{Supplementary movies: \url{https://rishabh-tifr.github.io/cyclic-denoising/movies}}

This procedure reveals a yielding-like stability spectrum. At low amplitudes, the system cannot explore much and the trajectories collapse into trivial absorbing states. These are either trivial fixed points consisting of featureless (near-monochromatic) images, or limit cycles consisting of simple Turing-like oscillating patterns. At intermediate amplitudes, some exploration of the landscape becomes possible, and the dynamics uncover simple memorized artifacts such as logos and web templates. At higher amplitudes, the dynamics can escape shallow basins, explore larger regions of the landscape and isolate even deeper attractors that persist for hundreds to thousands of cycles. We show that many of these attractors are not generic samples: they correspond to memorized training images and repeated web-crawl artifacts.

This yields a prompt-free extraction attack driven by the model's own dynamics, requiring only sampler-level control and no prior knowledge of, or access to, the training data. The signature of memorization is dynamical: when a stroboscopic trajectory locks into a basin and persists there for hundreds to thousands of cycles, the dynamics itself flags the state as a memorization candidate. We do not generate independent samples and then search for memorized ones after the fact; long-lived absorption determines which states are inspected. Furthermore, the protocol is agnostic to the initial condition: cycling can start from a generic image, a model-generated sample, or pure noise. We demonstrate the phenomenon in Stable Diffusion v1.4 and in a pixel-space DDPM trained on CIFAR-10. In both settings, memorized data appears not as isolated rare samples, but as dynamically stable attractors exposed by sustained cyclic perturbation.

Our results suggest a new dynamical route for probing memorization in diffusion models: training examples that persist as deep basins in the generative landscape, and are rarely encountered by standard sampling, can nevertheless be accessed through cyclic dynamics. Cyclic denoising therefore provides both a physics-inspired dynamical probe of diffusion-model generative landscapes and a practical tool for memorization auditing.

Our main contributions are:
\begin{itemize}[leftmargin=*,itemsep=1pt,topsep=2pt,parsep=0pt]
    \item \textbf{Cyclic denoising as a dynamical probe of diffusion-model generative landscapes.}
    We introduce cyclic denoising, a repeated forward--reverse diffusion protocol for traversing the generative landscape. By feeding the output of one cycle into the next, the protocol turns sampling into a stroboscopic dynamical system controlled by a cycling amplitude $\gamma$. Inspired by random organization and mechanical annealing in driven disordered systems, it probes the stability structure of the learned generative landscape rather than drawing independent samples from it.

    \item \textbf{Absorbing states, limit cycles, basin hopping, and ultrastable attractors.} We show that diffusion models exhibit rich long-time dynamics under repeated noising--denoising cycles. At low amplitudes, trajectories can collapse into trivial absorbing fixed points or simple limit cycles; at larger amplitudes, the dynamics become intermittent, with long residence times in attractor basins separated by transient exploratory hops between basins. Some basins confine trajectories for hundreds to thousands of cycles, and the deepest attractors are ultrastable, regenerating after severe corruption and persisting under sustained cyclic perturbation.

    \item \textbf{A yielding-like transition and amplitude-dependent stability spectrum.}
    Varying the cycling amplitude $\gamma$ reveals a dynamical transition from non-yielded low-amplitude behavior to intermittent exploration as the amplitude increases, with trajectories hopping between basins and becoming temporarily trapped in long-lived attractors. 
    The recovered attractor set depends systematically on $\gamma$: low amplitudes produce trivial fixed points or limit cycles, intermediate amplitudes recover simple logos and web-crawl artifacts, and higher amplitudes isolate deeper, richer memorized images. Thus, $\gamma$ acts as a stability filter, selecting different subsets of the model's attractor set.

    \item \textbf{A prompt-free, training-data-agnostic extraction attack.}
    We show that many non-trivial attractors correspond to memorized training images, yielding an extraction attack that requires only sampler-level control---the ability to partially noise a sample to an intermediate diffusion timestep and denoise it back. The attack requires no prior knowledge of the training data, captions, or prompts, and uses no gradients, weight inspection, post-hoc clustering, or membership inference to propose candidates: memorized candidates appear directly as persistent states in the cyclic dynamics.

      \item \textbf{Stability as a memorization diagnostic for prompted models.}
      We extend cyclic denoising to prompt-conditioned sampling and measure stability after the prompt is removed. Post-removal stability correlates positively with
  memorization: prompt-stabilized concept basins decorrelate fastest, while genuinely memorized training images span a wide range of stabilities, with the deepest remaining
  ultrastable. This makes the post-removal decorrelation time a $\gamma$-dependent dynamical test for memorization of candidate prompt--image pairs.
\end{itemize}

\begin{figure}[htpb]
  \centering
  \includegraphics[width=1.0\linewidth]{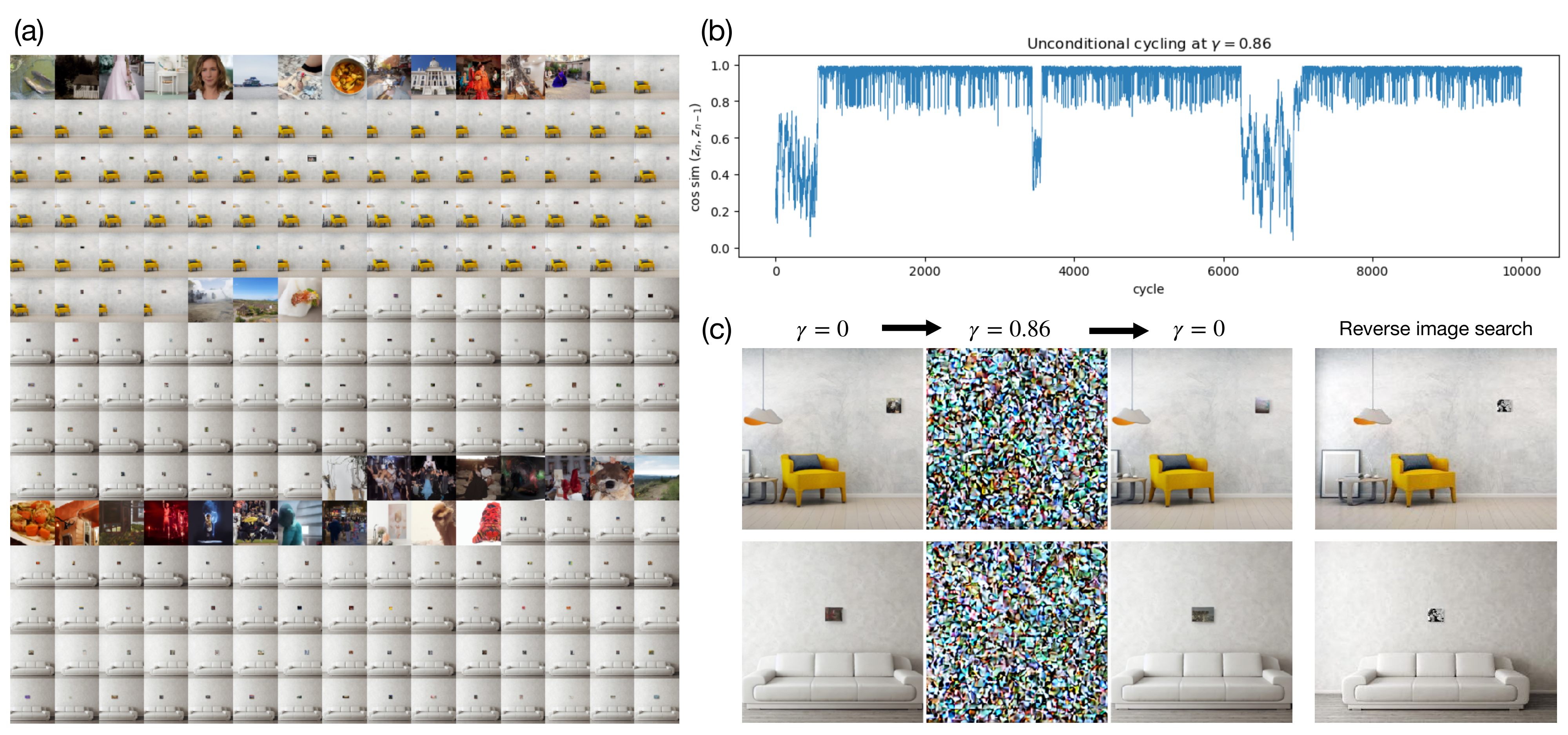}
  \caption{\textbf{Unconditional cyclic denoising drives latents toward attractors in the Stable Diffusion v1.4 landscape.} (a) Decoded snapshots from a $10^4$-cycle unconditional cycling trajectory at $\gamma=0.86$, starting from a single ImageNet test image; the $15\times15$ grid shows every 44th cycle, in reading order, starting from the top-left. After an initial transient (snapshots change cycle-to-cycle), the dynamics lock into the first attractor — a room with a yellow chair — that persists for over 2000 cycles. A brief transient carries the
  trajectory into a second attractor (white-couch scene, paintings on a gray wall), which survives for a comparable number of cycles before another, longer transient returns the trajectory to the same second basin. (b) Cosine similarity between consecutive latents $z_n$ and $z_{n-1}$ along the trajectory. Plateaus near unity correspond to residence within a basin; sharp drops mark inter-basin transitions,
  matching the qualitative changes in (a). Cosine similarities throughout this work are measured stroboscopically — at the end of each completed cycle. (c) A single forward--reverse cycle, $\gamma=0 \rightarrow 0.86 \rightarrow 0$, for an example latent drawn from each attractor. The room layout is regenerated after near-total corruption, while fine details such as wall artwork fluctuate stochastically---the signature of an attracting basin in which the dominant scene template is locked while peripheral content remains diffusive. Right column: reverse image search retrieves near-duplicate web images for both attractors, corresponding to e-commerce template scenes in which different artworks are repeatedly composited onto the same room layout. These templates recur across many vendor websites, making them very likely to be duplicated in the web-scale training crawl. Independently, \citet{somepalli2023understandingmitigatingcopyingdiffusion} flag these same two scenes (their Fig.~2) among the most heavily replicated Stable Diffusion v1.4 generations and trace them to duplicated LAION training images---training-set-grounded confirmation that the attractors recovered here by cycling alone, with no training-set access, are genuine memorized content. The two scenes also appear under similar product captions referring to the displayed artwork rather than the room itself, suggesting a possible mechanism for their proximity in the learned generative landscape. Cyclic denoising thus exposes these memorized templates from the model's dynamics alone. The URLs for the reverse image search are listed in the Appendix (Table~\ref{tab:Supplementary_Table1}). }
  \label{fig:main_Figure1}
\end{figure}

\section{Related work}

\subsection{Dynamical views of diffusion and generative models}

Recent work has begun to treat generative models not only as samplers, but also as dynamical systems whose trajectories reveal structure in the learned distribution. For diffusion models, the forward noising process and learned reverse denoising process provide a natural perturb-and-reconstruct experiment. \citet{wyartPNAS2025} showed that a single forward--backward diffusion setup, or a single u-turn, can probe the hierarchical organization of data: in a hierarchical generative model, the probability of reconstructing high-level features such as class identity drops sharply beyond a threshold diffusion time, while lower-level details evolve smoothly across the whole diffusion process. Their results frame diffusion time as a scale-resolved probe of learned structure, with different levels of the data hierarchy being modified at different noise amplitudes. Our work builds on this forward--backward viewpoint but changes the experiment qualitatively. Rather than performing a single noising--denoising pass, we iterate the partial forward--reverse map for thousands of cycles. This turns diffusion sampling into a stroboscopic dynamical system. The resulting long-time dynamics reveal phenomena that are invisible to one-shot reconstruction experiments: absorbing states, limit cycles, basin hopping, cyclic yielding-like transitions, and ultrastable attractors. Thus, while prior forward--backward studies ask which features survive one corruption--reconstruction pass, cyclic denoising asks which states are reached and remain stable under repeated perturbation.

A complementary line of work studies memorization through the \emph{training} dynamics of diffusion models. \citet{biroli2025} identify two characteristic timescales over the course of training: a generalization time $\tau_{\mathrm{gen}}$, at which the model begins to produce high-quality samples, and a later memorization time $\tau_{\mathrm{mem}}$, beyond which memorization emerges. They show that $\tau_{\mathrm{mem}}$ grows approximately linearly with the training-set size $n$, while $\tau_{\mathrm{gen}}$ remains approximately constant, opening a generalization window $[\tau_{\mathrm{gen}},\tau_{\mathrm{mem}}]$ that widens with $n$. Stopping training in this window lets even highly overparameterized models generalize before eventually memorizing, a form of implicit dynamical regularization. 
Our focus is orthogonal: we hold a trained model fixed and ask whether memorized examples persist in its post-training sampling dynamics. Cyclic denoising probes this post-training stability structure, revealing memorized images as long-lived attractors even when they are rarely encountered by standard sampling.

A closely related dynamical perspective is the work of \citet{fumero2026}, who interpret autoencoder-based neural models as defining latent vector fields by iterating the encode--decode map. In their setting, an autoencoder induces a discrete-time map in latent space, and the resulting trajectories and attractors can be used to analyze generalization, memorization, prior knowledge encoded in the network, and out-of-distribution behavior. This is conceptually aligned with our view that model dynamics can expose learned structure. However, the extension to diffusion models is not direct. If one fully noises a diffusion sample to the terminal time before denoising, the previous state is erased and the reverse process becomes a fresh draw from the learned generative distribution. In this full-noising limit, iterating forward--reverse diffusion does not produce a meaningful trajectory with memory; it degenerates into repeated resampling. Cyclic denoising operates instead in the partial-memory regime with noise level $\gamma<1$. Each cycle noises the current state only to an intermediate amplitude and then denoises it back, so the intermediate state retains partial information about the previous sample.
Crucially, diffusion models provide a natural control knob: by varying $\gamma$, we tune how strongly the current state is perturbed before reconstruction. Inspired by cyclic yielding in disordered systems, where changing the drive amplitude qualitatively changes the states explored by the dynamics, we use $\gamma$ to probe the stability hierarchy of the learned generative landscape. Low amplitudes reveal shallow or trivial absorbing states and limit cycles; intermediate amplitudes uncover simple memorized artifacts such as logos and web templates; and higher amplitudes isolate deeper attractors corresponding to richer memorized images. Thus, varying $\gamma$ is central to the method: it turns diffusion time (or equivalently $\gamma$) into a stability-resolved probe of memorization.

\subsection{Energy landscapes and associative-memory views of diffusion models}

A complementary line of work interprets diffusion models through the lens of associative memory.
\citet{DenseAM_Krotov2025} recasts diffusion training as memory encoding and generation as memory retrieval, drawing an analogy to Hopfield and Dense Associative Memory systems in which stored patterns correspond to attractors of an energy landscape.
Across the memorization--generalization transition, they distinguish memorized samples, emergent \emph{spurious} attractors absent from the training data, and generalized samples, and characterize these states using basin-volume estimates and energy-curvature spectra.
This provides strong support for an attractor-based view of diffusion memorization: memorized examples can behave as deep, large-basin states of the learned landscape. Our contribution is complementary. The associative-memory analysis identifies and characterizes attractors, but does not by itself specify a cyclic dynamical route for reaching or discovering memorized basins in a fixed sampler.
We import that route from the physics of periodically driven disordered systems---random organization and yielding---where finite-amplitude cycling is the canonical probe of a rugged landscape. Their basin probes, by contrast, start from candidate states already in hand and test whether these reconstruct after a single perturbation; cyclic denoising instead drives the sampler repeatedly from arbitrary initial conditions and lets attractors emerge from the dynamics themselves. The cycling amplitude is the control parameter: at small amplitude the dynamics remain non-yielded, settling into trivial absorbing states or simple limit cycles, while at larger amplitude they hop intermittently between basins and dwell in long-lived attractors, among them memorized images.

\subsection{Training-data extraction from diffusion models}

Training-data extraction attacks ask whether a trained generative model can be made to reproduce examples from its training set. In diffusion models, this has been studied as both a privacy risk and a copyright/compliance problem. Using retrieval-based comparisons between generated samples and training images, \citet{somepalli2022diffusionartdigitalforgery} showed that text-to-image diffusion models, including Stable Diffusion, can directly replicate training content. Follow-up work showed that such copying is not explained by image duplication alone: the model's text conditioning plays a comparably important role, with replication common in text-conditional models but much less frequent in unconditional ones~\citep{somepalli2023understandingmitigatingcopyingdiffusion}.

The canonical extraction attack on diffusion models was introduced by~\citet{carlini2023extractingtrainingdatadiffusion}. 
Their text-to-image attack follows a generate-and-filter paradigm: generate many samples, identify candidate near-duplicates, and verify memorization using similarity or membership-inference-style tests. 
For Stable Diffusion, they target captions associated with highly duplicated training examples: they select the $350{,}000$ most-duplicated examples, generate $500$ samples per caption, and then filter the resulting $175$ million generations for near-identical cliques. A prompt is flagged when at least $10$ of its $500$ generations collapse to near-duplicates under a patch-based image distance, and candidates are then verified against the training set. 
This large-scale pipeline yields $94$ confirmed Stable Diffusion extractions under their strict $(\ell_2,0.15)$ criterion, rising to $109$ near-copies under manual inspection. For unconditional CIFAR-10 diffusion models, where prompts are unavailable, \citet{carlini2023extractingtrainingdatadiffusion} use a different controlled setup: they train $16$ diffusion models, each on a random half of the CIFAR-10 training set, generate $2^{20}$ unconditional samples ($\approx 1.05$ million) in total, and identify memorized examples by direct comparison to the training set using calibrated nearest-neighbor distances. 
This recovers $1{,}280$ unique CIFAR-10 training images, but relies on a known, searchable training set and multiple models trained on different data subsets. Thus, while prior work demonstrated that diffusion models can leak training data, these pipelines rely on auxiliary information or procedures such as prompts, captions, training-set access, membership-inference criteria, multi-model training, or large-scale post-hoc filtering. In contrast, cyclic denoising proposes candidates \emph{from the model's own long-time dynamics}. In both Stable Diffusion and CIFAR-10, we operate on a single fixed checkpoint trained on the full dataset and surface candidates from the dynamics alone---none of these signals required.

\citet{webster2023reproducibleextractiontrainingimages} take a different route from large-scale generate-and-filter attacks by exploiting one-step synthesis behavior in text-to-image diffusion models. 
They observe that some prompts corresponding to memorized images can reproduce them near-verbatim after only a single denoising step, and turn this into fast scoring rules for candidate captions: a denoising confidence score in the white-box setting and an edge-consistency score in the black-box setting. 
This reduces the number of network evaluations by orders of magnitude relative to sampling hundreds of full trajectories per prompt, but the attack remains driven by known captions---candidates are generated from captions, scored, and then labeled by matching to the corresponding training image, or, for template verbatims, by retrieval and masking against the training set. 
In contrast, cyclic denoising neither scores captions nor tests one-step reconstruction; it surfaces candidates as attractors of the long-time cyclic dynamics, without any conditioning.

Unconditional diffusion models are harder to attack because there is no prompt with which to steer the model toward memorized samples. 
SIDE addresses this setting with \emph{surrogate conditioning}: it generates synthetic samples from the target model, clusters them in a pretrained feature space to form data-driven pseudo-labels, and uses these surrogate labels to guide extraction from the original model~\citep{chen2025sidesurrogateconditionaldata}. In practice, this guidance is implemented through additional learned machinery, such as a time-dependent classifier or LoRA-based fine-tuning. Cyclic denoising instead uses the original sampler dynamics alone. 
To propose candidates, it requires only sampler-level control---the ability to partially noise a sample to an intermediate diffusion timestep and denoise it back---and none of SIDE's surrogate-conditioning machinery: no pretrained feature extractor, no clustering of generated samples, no auxiliary classifier or LoRA fine-tuning, and no classifier-guidance gradients.

Across these attacks, candidates are found by generating large pools of samples---from known or suspected training captions in the conditional case, and unconditionally where no captions are available---that are then clustered, scored, or verified after the fact. Cyclic denoising instead follows a single correlated trajectory: memorized candidates reveal themselves as attractors, long-lived basins, or high-similarity plateaus under repeated noising--denoising cycles.

\section{Methods}
We use two open-weight diffusion models without modifying their architectures or weights: Stable Diffusion v1.4, a latent diffusion model operating in a $4\times64\times64$ VAE latent space~\citep{Rombach2022} (\url{https://huggingface.co/CompVis/stable-diffusion-v1-4}), and an unconditional pixel-space DDPM trained on CIFAR-10~\citep{ho2020DDPM} (\url{https://huggingface.co/google/ddpm-cifar10-32}). Stable Diffusion is run unconditionally (empty prompt, guidance scale $0$) with $N=50$ inference steps, and the CIFAR-10 DDPM with $N=250$ inference steps.

\begin{figure}[htpb]
  \centering
  \includegraphics[width=1.0\linewidth]{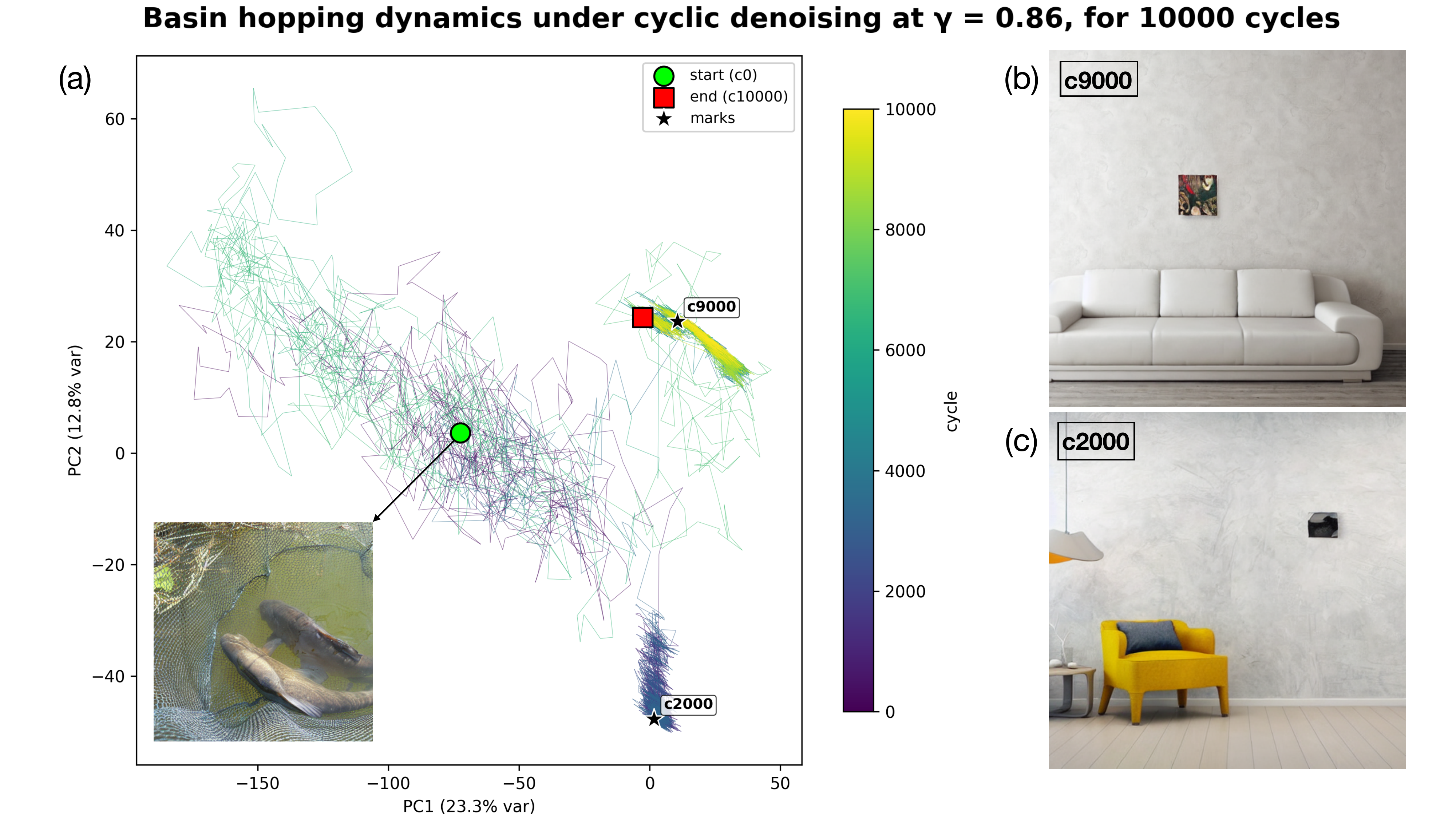}
  \caption{\textbf{Basin hopping in latent space under unconditional cyclic denoising
  in Stable Diffusion v1.4.}
  The same trajectory shown in Fig.~\ref{fig:main_Figure1}, here visualized as a two-dimensional PCA
  projection of its latent trajectory; the top two components are fit on this
  trajectory alone, and line color encodes cycle number. A single ImageNet
  test-set image (green circle, cycle~0, inset) is cycled through Stable
  Diffusion v1.4 at~$\gamma=0.86$ for 10{,}000 cycles (unconditional cycling).
  After a short transient, the trajectory is captured by a first attractor in
  which it dwells for many cycles, before escaping and hopping to a second,
  well-separated attractor; it then briefly leaves this second basin but is
  recaptured by it, ending at cycle~10{,}000 (red square). Black stars mark two
  individual states along the trajectory --- cycle~2000 (\texttt{c2000}), in the
  first basin, and cycle~9000 (\texttt{c9000}), in the second; panels (c) and (b)
  show these two states decoded to pixels, as representative examples from each
  basin. Rather than decorrelating, the dynamics dwell for long stretches in deep absorbing states and hop between them, revealing the model's ultrastable memories. See Supplementary Movie~1 for the complete trajectory.}
  \label{fig:main_Figure2}
\end{figure}

Starting from a state $z_n$, one cycle of \emph{cyclic denoising} consists of noising the state up to an intermediate noise level and then denoising it fully back to a clean state; the output becomes the input to the next cycle, defining a stroboscopic trajectory observed at the end of each noising--denoising cycle. The control parameter is the cycling amplitude $\gamma\in[0,1]$, the fraction of the inference trajectory traversed per cycle, which sets how much noise is added before denoising. Small $\gamma$ adds little noise and gives a near-identity perturbation, while $\gamma=1$ corresponds to near-complete noising followed by a full denoising pass; in every case we denoise all the way back to a clean state, so $\gamma$ controls only how high each cycle climbs. We use the DDPM sampler throughout, with fresh Gaussian noise drawn both when noising and during denoising, so that cycling is a diffusion analogue of stochastic cyclic driving rather than a deterministic map.

For Stable Diffusion, we use three initialization ensembles: ImageNet validation images encoded into the VAE latent space, model-generated latents, and i.i.d. Gaussian latents. For CIFAR-10, we initialize from randomly sampled CIFAR-10 training images. We run $10{,}000$ cycles per trajectory while sweeping $\gamma$ over a range of amplitudes; Stable Diffusion uses 10 seeds per initialization ensemble and CIFAR-10 uses 50 training-image seeds, with the seed set held fixed across all amplitudes in both cases. At each cycle, we compute the cosine similarity between consecutive stroboscopic states (in the VAE latent for Stable Diffusion and in pixel space for CIFAR-10), $\cos(z_n,z_{n-1})$. The steady-state similarity reported in yielding curves is the average over the final $1000$ cycles of each trajectory, followed by an average across seeds. Absorbing episodes are identified by long plateaus of near-unity stroboscopic similarity, and their decoded states are inspected as candidate attractors. Stable Diffusion attractors are verified post hoc using public reverse-image search. CIFAR-10 attractors are verified post hoc by cross-checking against memorized examples reported by \citet{carlini2023extractingtrainingdatadiffusion}. These verification steps are not used to generate candidates. For prompted experiments, we use the same cyclic protocol with classifier-free guidance. Trajectories are first cycled with a candidate prompt; in prompt-removal experiments, we then continue cycling unconditionally, tracking the cosine $\cos(z_t,z_0)$ between each cycle and the image absorbed at the moment of removal to measure how long the recovered state remains stable without conditioning.

\begin{figure}[tpb]
  \centering
  \includegraphics[width=0.75\linewidth]{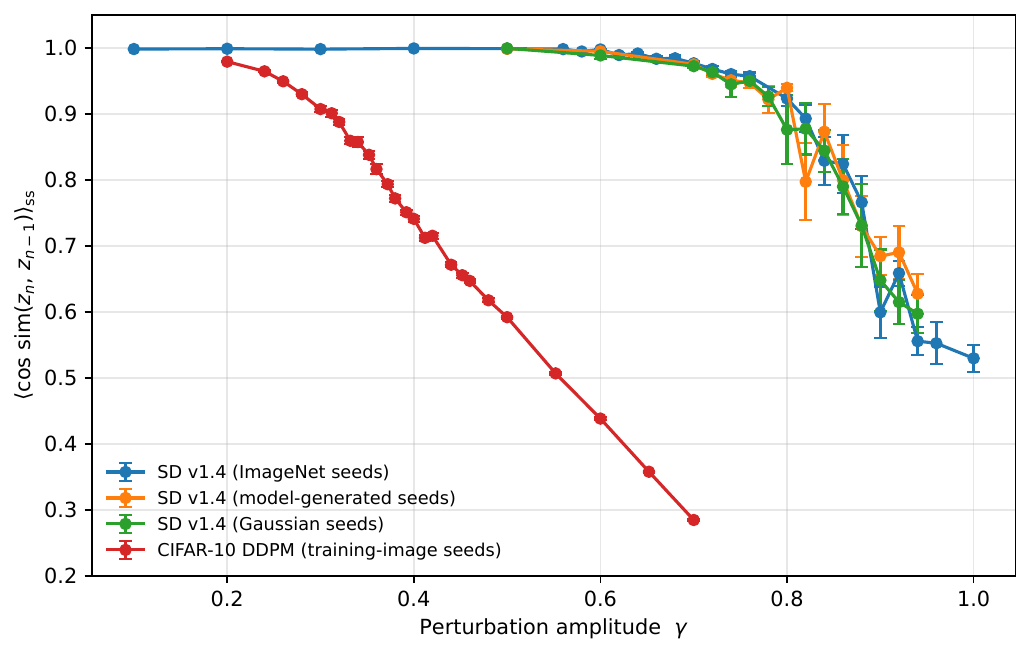}
\caption{\textbf{Yielding diagram for cyclic denoising.}
  Steady-state stroboscopic similarity $\langle \cos(z_n,z_{n-1})\rangle_{\mathrm{ss}}$ between consecutive cycles versus the perturbation (cycling) amplitude $\gamma$.
  Each trajectory is run for $10{,}000$ cycles; for each seed we average the similarity over the final $1000$ cycles, then report the mean across seeds with $\pm$SEM error
  bars.
  Stable Diffusion v1.4 is shown for ImageNet, model-generated, and Gaussian initializations ($10$ seeds per $\gamma$ each); the CIFAR-10 DDPM is initialized from CIFAR-10
  training images ($50$ seeds per $\gamma$).
  At low $\gamma$ the dynamics are absorbing --- consecutive cycles return to essentially the same state ($\langle\cos\rangle_{\mathrm{ss}}\approx1$), the signature of the
  fixed points and limit cycles of the stroboscopic map.
  Beyond a model-dependent critical amplitude the similarity drops sharply: a yielding-like transition in which cycling drives rearrangements, basin hopping, and exploration
  of the landscape.
  The three Stable Diffusion initialization ensembles collapse onto a common master curve, indicating that the transition is a property of the learned generative landscape
  rather than of the initial condition.
  The CIFAR-10 DDPM yields at markedly lower $\gamma$ than Stable Diffusion; we attribute this partly to the models' different noise schedules, under which the step-fraction
  $\gamma$ maps nonlinearly --- and differently --- onto signal-to-noise ratio, so $\gamma$ is comparable only within a model and not across the two. The qualitative
  absorbing-to-exploratory transition is nonetheless shared across latent- and pixel-space diffusion. Also note that even at $\gamma=1$ the similarity does not fall to zero: Stable Diffusion's noise schedule has a non-zero terminal SNR, so a small fraction of the previous latent
  is still carried over each cycle rather than a true reset to pure noise.}
  \label{fig:main_Figure3}
\end{figure}

\textbf{Compute resources.}
Each cycling run is an independent single-GPU job on one NVIDIA L40S (48 GB VRAM); the only parallelism is running separate amplitudes concurrently, with no multi-GPU or distributed execution within a run. Per-run time scales roughly linearly with the cycling amplitude, since a cycle at amplitude $\gamma$ runs proportionally more denoising steps: across the swept amplitudes, a $10{,}000$-cycle Stable Diffusion run (10 seeds batched through the UNet) ranges from a few hours at low $\gamma$ to ten or eleven hours near $\gamma=1$, with an intermediate amplitude ($\gamma\approx0.7$) taking about eight hours. The CIFAR-10 DDPM ($N=250$, 50 seeds) follows the same linear-in-$\gamma$ scaling. Prompt-conditioned cycling with classifier-free guidance doubles the per-step UNet cost, since each step requires conditional and unconditional predictions, but uses only 5 seeds, keeping wall-clock time comparable. The total compute footprint of all results in this paper is within $1000$ L40S-GPU-hours, substantially below large-scale caption-based extraction pipelines that rely on generating and filtering millions of images. We stress that this budget quantifies the full dynamical phenomenology rather than extraction alone: it spans the entire amplitude range needed to map the yielding diagram (Fig.~\ref{fig:main_Figure3}), including low-$\gamma$ runs where the dynamics settle into trivial absorbing states or simple limit cycles and do not produce non-trivial memorized candidates in our runs. Extraction does not require this full coverage; an attack could focus on the intermediate-to-large amplitudes where memorized attractors, basin hopping, and long residence times occur. Thus, the compute required for extraction alone can be substantially lower than the totals reported here.

\begin{figure}[tpb]
  \centering
  \includegraphics[width=1.0\linewidth]{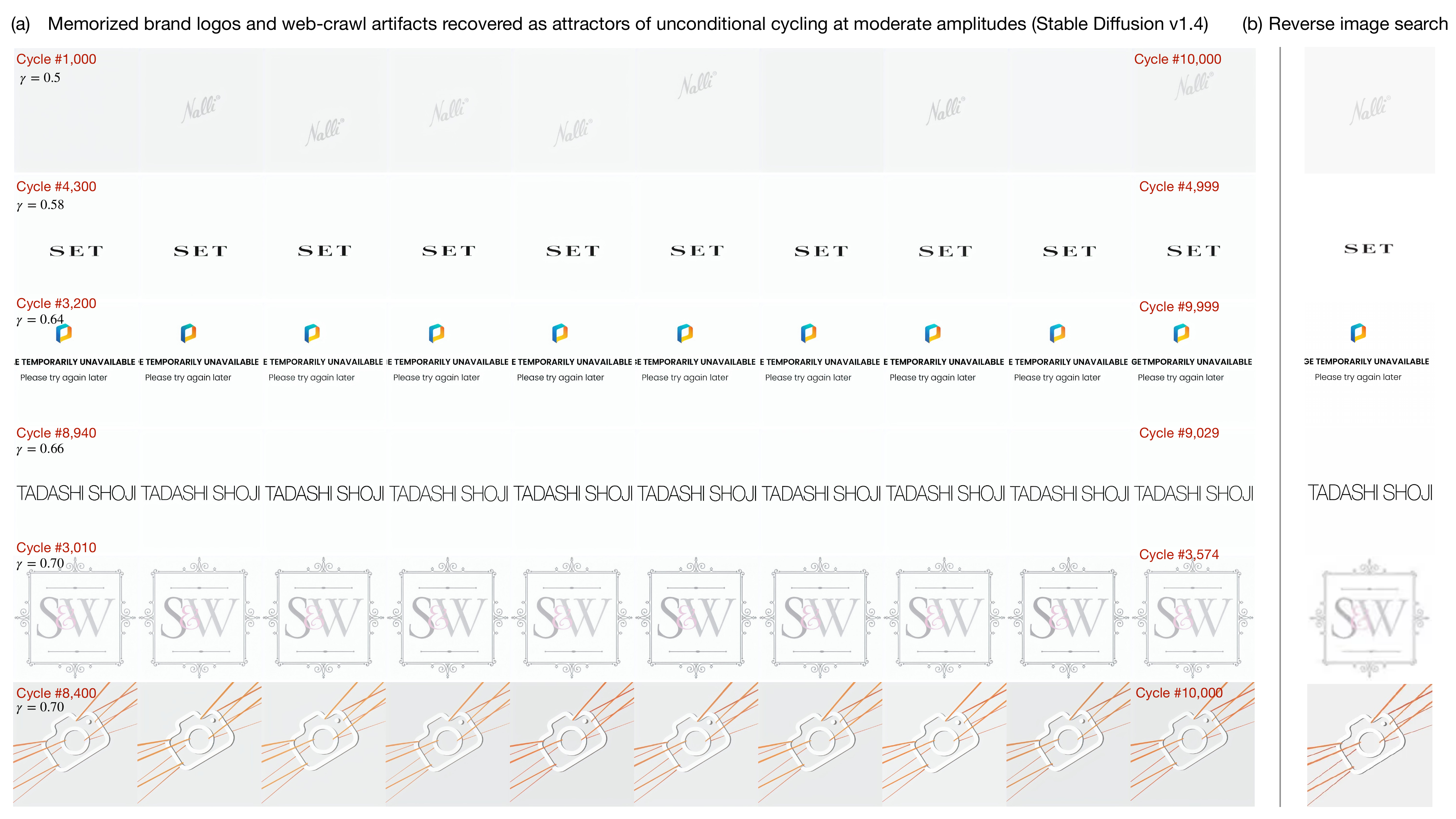}
 \caption{\textbf{Moderate-amplitude cycling recovers memorized logos and web-crawl artifacts.}
(a) Selected basins recovered by unconditional cyclic denoising in Stable Diffusion v1.4 at moderate noise amplitudes. Each row shows ten equispaced stroboscopic snapshots from a single cycling trajectory during the residence time of one basin, with the corresponding cycle range and cycling amplitude indicated. At very low $\gamma$, the dynamics collapse to trivial absorbing states---saturated patterns, monotonic fills, and simple geometric shapes---as the trajectory becomes trapped after iterating on local details, lacking the perturbation amplitude needed to rearrange the latent at larger scales, see Supplementary Fig.~\ref{fig:supplementary_Figure1}. In the intermediate regime shown here, the perturbation is large enough to escape these trivial fixed points; the resulting attractors are simple, highly repeated visual templates, including brand logos, product-page placeholders, and web-crawl artifacts.
(b) Reverse image search retrieves close web matches for each recovered attractor, indicating that these states correspond to memorized image templates rather than generic samples. These attractors are recovered across multiple random seeds and nearby values of $\gamma$, but typically become transient at larger amplitudes, placing them at intermediate depth in the basin-stability spectrum: more structured than trivial low-amplitude absorbing states, but less stable than the deep memorized scenes that survive high-amplitude cycling.}
  \label{fig:main_Figure4}
\end{figure}

\begin{figure}[tpb]
  \centering
  \includegraphics[width=1.0\linewidth]{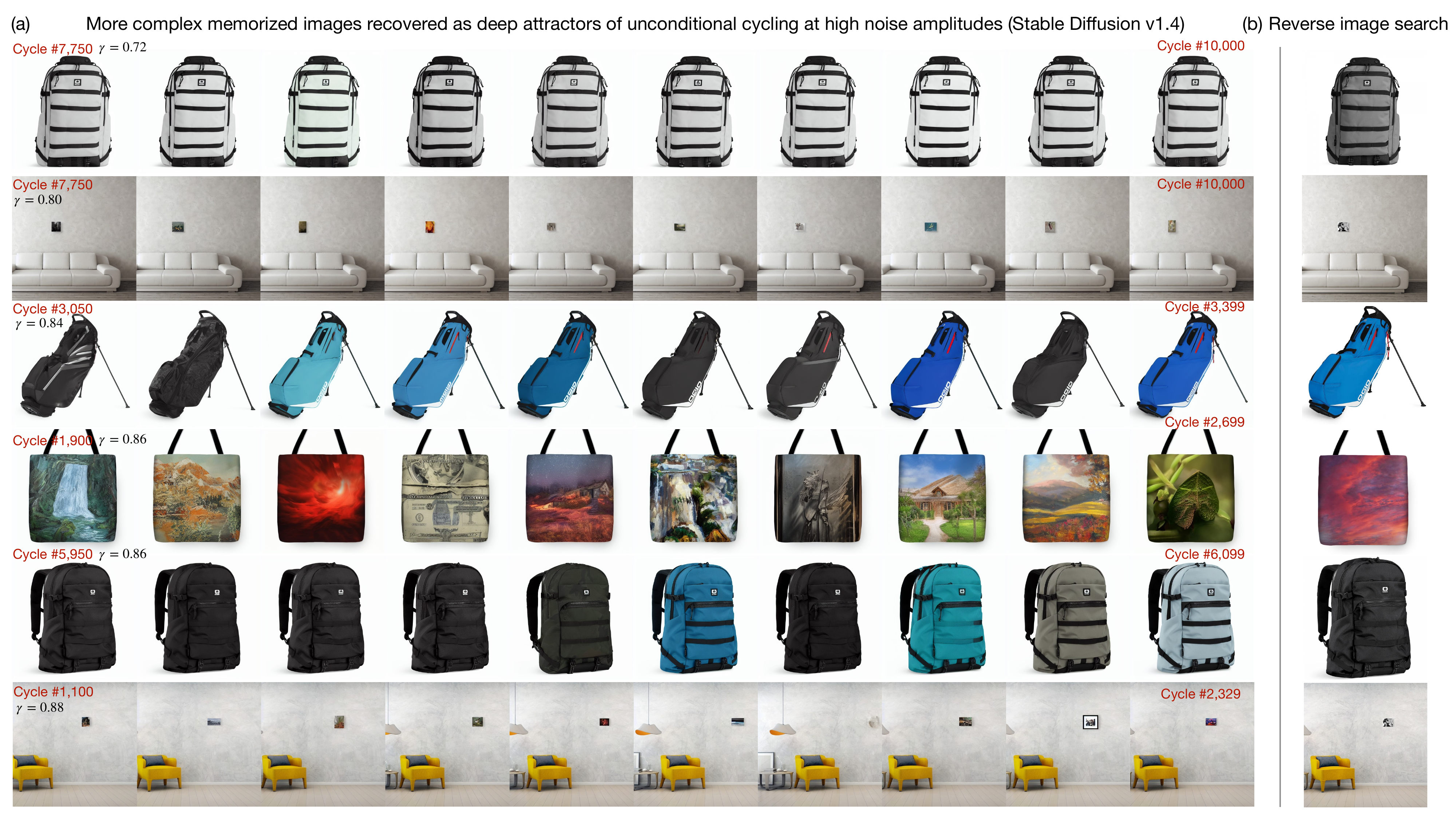}
\caption{\textbf{High-amplitude cycling isolates deep memorized attractors.}
(a) Memorized images recovered by unconditional cyclic denoising in Stable Diffusion v1.4 at high noise amplitudes. Each row shows ten equispaced stroboscopic snapshots from a single cycling trajectory during the residence time of one attractor, with the cycle range and $\gamma$ indicated. As in random organization, cyclic forcing drives the system toward states that resist change under the applied drive. At higher amplitudes, the trajectory escapes shallow basins and explores more of the learned distribution, revealing only deeper attractors as long-lived states. The recovered attractors include richer memorized images than the intermediate-amplitude logos and web artifacts in Fig.~\ref{fig:main_Figure4}, including product photographs, stock-style object renderings, and recurring room templates.
(b) Reverse image search retrieves close web matches for the recovered attractors, indicating that these states correspond to memorized training images or templates. These deep attractors recur across random seeds, initial conditions, and nearby values of $\gamma$, with some basins---especially the living room (with the white sofa and yellow chair) scenes---capturing many trajectories for long residence times. Including Fig.~\ref{fig:main_Figure4}, the recovered attractor set forms a dynamical fingerprint of Stable Diffusion v1.4: a model-specific signature of memorized training content exposed by unconditional cycling alone. Representative reverse-image-search source URLs are listed in the Appendix (Table~\ref{tab:Supplementary_Table1}).}
  \label{fig:main_Figure5}
\end{figure}

\begin{figure}[tpb]
  \centering
  \includegraphics[width=1.0\linewidth]{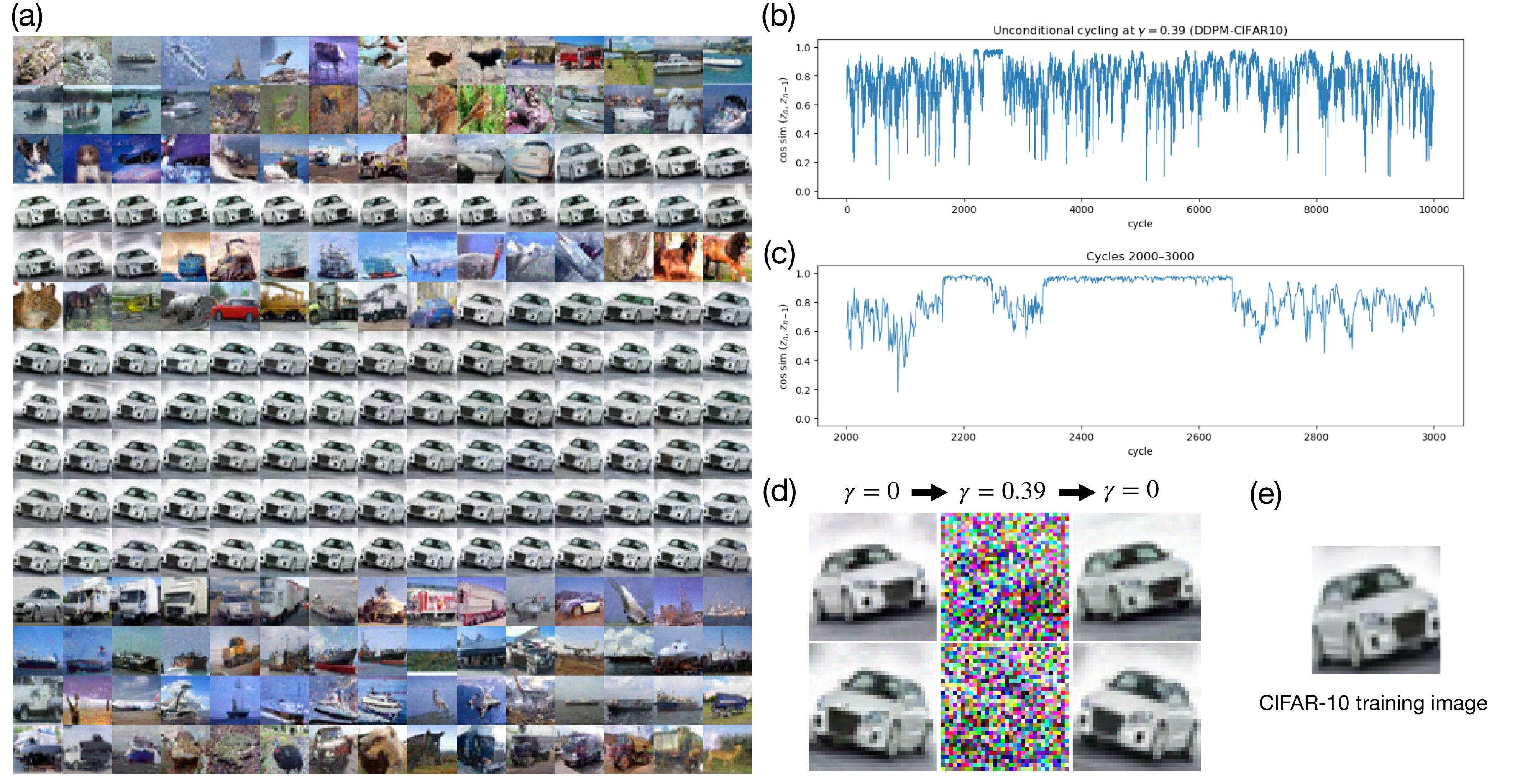}
    \caption{\textbf{Unconditional cyclic denoising drives CIFAR-10 DDPM samples toward memorized attractors.} (a) Snapshots from a $10^4$-cycle unconditional cycling trajectory in a pixel-space DDPM trained on CIFAR-10, initialized from a CIFAR-10 training image. The $15\times15$ grid shows every fourth cycle from cycles 2000--2896. After a short transient, the trajectory locks into a car-like attractor, briefly leaves it, and then enters a nearby mirror-image attractor, where it persists for many
  cycles, before ultimately escaping again. (b) Cosine similarity between consecutive cycle-end samples $z_n$ and $z_{n-1}$ along the full trajectory, measured stroboscopically at the end of each completed cycle. (c) Zooming into cycles 2000--3000 reveals the plateau--drop--plateau structure corresponding to residence in one basin, transient hopping, and residence in the mirror-related basin. The paired attractors suggest that horizontal-flip augmentation during training may encode symmetry-related memorized states as nearby basins. This resembles two-level systems in disordered solids, where two nearly degenerate configurations are separated by a small barrier and the system can switch between them. (d) A single forward--reverse cycle, $\gamma=0 \rightarrow 0.39 \rightarrow 0$, for representative samples from the two basins. (e) Matching CIFAR-10 training image reported in~\citep{carlini2023extractingtrainingdatadiffusion}. The memorized candidate appears directly as persistent high-similarity plateaus in the cyclic trajectory. Unlike prior CIFAR-10 extraction, which required training-set lookup to identify memorized samples, cyclic denoising recovers the candidate from the model dynamics alone. See Supplementary Fig.~\ref{fig:Supplementary_Figure5} for additional recovered CIFAR-10 attractors.}
  \label{fig:main_Figure6}
\end{figure}

  \begin{figure}[tpb]
    \centering
    \includegraphics[width=1.0\linewidth]{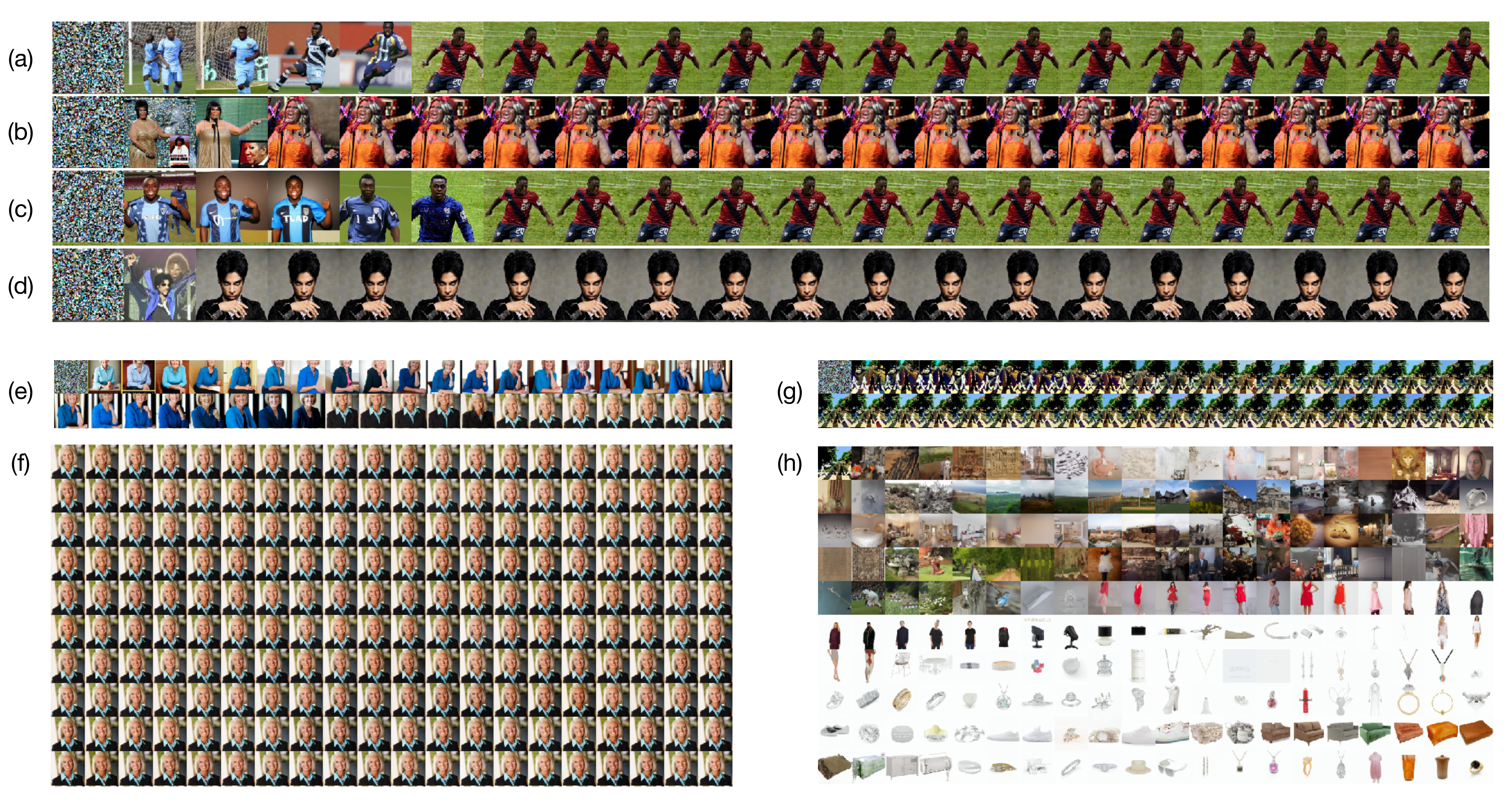}
      \caption{\textbf{Stability of prompt-conditioned image attractors under cyclic denoising.} (a--d) Prompt-conditioned cycling in Stable Diffusion v1.4 at $\gamma=0.78$ using classifier-free guidance and prompts previously associated with memorized samples~\citep{carlini2023extractingtrainingdatadiffusion,webster2023reproducibleextractiontrainingimages}. Rows show: (a) ``Freddy Adu Signs For Yet Another Club You Probably Don't Know''; (b) ``Aretha Franklin Files \$10 Million Suit Over Patti LaBelle Fight Story On Satire Website''; (c) the same Freddy Adu prompt as in (a), initialized from a different Gaussian seed; and (d) ``Prince Reunites With Warner Brothers, Plans New Album.'' Each trajectory rapidly converges to the corresponding known memorized image and remains absorbed; the convergence of (a) and (c) to the same image from different seeds demonstrates seed-independent attraction. (e,f) Stability of the ``Ann Graham Lotz'' memory, following one example trajectory (one of five Gaussian seeds). (e) With the prompt present, the trajectory converges to the memorized image and remains locked for the full $5000$-cycle run; all five seeds converge to the same image, with convergence times that vary across seeds. (f) Continuing the same trajectory unconditionally after prompt removal, it stays locked in the same image for the full additional $1000$ cycles we ran---still absorbed at the end rather than decorrelating; of the five seeds, four remain locked, indicating that the prompt steered the system to a state that is itself ultrastable under the unconditional dynamics. (g,h) A prompt-stabilized concept attractor---stable under the prompt but not unconditionally absorbing---again following one example trajectory. (g) Cycling with the guessed prompt ``Abbey Road album cover''---not known to correspond to any training image---the trajectory reaches a steady prompted state that preserves the iconic zebra-crossing composition without collapsing to a single image, a crude rather than verbatim rendering (Supplementary Fig.~\ref{fig:Supplementary_Figure6}). (h) After prompt removal it decorrelates within a few cycles, as do all five seeds---the fastest decay among the prompts tested (Supplementary Fig.~\ref{fig:Supplementary_Figure7}).}
    \label{fig:main_figure7}
  \end{figure}
  
\section{Results}

\textbf{Unconditional cyclic denoising produces absorbing attractors.} Figure~\ref{fig:main_Figure1} shows the basic phenomenon in Stable Diffusion v1.4. A single unconditional cyclic trajectory does not produce independent samples; instead, it exhibits long-lived attractor residence, basin hopping, and partial absorption, where coarse scene structure remains stable while fine details fluctuate. Reverse image search confirms that the recovered attractors correspond to memorized web templates. Figure~\ref{fig:main_Figure2} visualizes the basin-hopping and transient-exploration phases in the PCA space, with clear clusters corresponding to the two memorized images. An animated version of this basin-hopping trajectory---its latent dynamics in PCA alongside the decoded images---is provided as Supplementary Movie~1.

\textbf{Different amplitudes expose different memories.} Figures~\ref{fig:main_Figure4} and~\ref{fig:main_Figure5}, along with Supplementary Fig.~\ref{fig:supplementary_Figure1}, show that the complexity of recovered attractors depends on $\gamma$. Intermediate amplitudes recover simple, highly repeated artifacts such as logos, product placeholders, and web-crawl remnants, whereas higher amplitudes isolate richer memorized images such as product photographs, object renderings, and recurring room templates. Supplementary Fig.~\ref{fig:Supplementary_Figure4} shows some additional examples of how individual trajectories traverse the landscape across intermediate-to-high amplitudes, sometimes collapsing into deep basins and, at others, wandering between semantically coherent shallow basins via inter-basin hops. At very high noise levels ($\gamma\to1$) successive cycles become essentially independent draws: memorized images may still recur---as they do across repeated independent generations in generate-and-filter attacks---but they no longer form the temporally contiguous dwelling episodes (connected runs of consecutive cycles within a single basin, colored coherently by cycle number) seen at intermediate $\gamma$. We also note that the same memorized images recur across different seeds, initialization ensembles, and nearby amplitudes (Supplementary Fig.~\ref{fig:Supplementary_Figure3}), so our recurrent attractor set forms a dynamical fingerprint of Stable Diffusion v1.4. Figure~\ref{fig:main_Figure3} quantifies this $\gamma$-dependence as a yielding diagram: at low amplitudes the dynamics are absorbing, returning to essentially the same state each cycle as the system settles onto a fixed point or, more rarely, a limit cycle that repeats with a fixed period in cycle number (Supplementary Fig.~\ref{fig:supplementary_Figure2}); beyond a critical amplitude the cycle-to-cycle return drops sharply as cycling induces rearrangements and basin hopping. This mirrors the yielding transition seen routinely in disordered solids under cyclic drive, where increasing the drive amplitude takes the system from a stuck phase---trapped in an absorbing state that returns to itself each cycle---to an (initial-state independent) fluidized phase that explores configuration space. In our cyclic denoising, trajectories in this exploratory regime are intermittently captured by deep attractors for hundreds to thousands of cycles before escaping and continuing to explore the landscape.

  \textbf{The phenomenon is not specific to latent diffusion.} Figure~\ref{fig:main_Figure6} shows the same absorbing-state dynamics in a pixel-space DDPM trained on
  CIFAR-10, confirming that these absorbing basins are not an artifact of the latent space or the VAE decoder. Starting from a CIFAR-10 training image and cycling at
  $\gamma=0.39$, the dynamics lock onto a memorized car as a persistent attractor, read off directly from the high-similarity plateau with no training-set lookup; the
  recovered image matches a CIFAR-10 example independently flagged as memorized by~\citet{carlini2023extractingtrainingdatadiffusion} (Fig.~\ref{fig:main_Figure6}(e)). After
  freezing on the car for many cycles, the trajectory passes through a brief transient into a nearby, horizontally mirrored variant, freezes there as well, and then escapes
  (Fig.~\ref{fig:main_Figure6}(c))---the same basin hopping between long-lived absorbing states we observe in the latent-space model (Figs.~\ref{fig:main_Figure1}
  and~\ref{fig:main_Figure2}). These nearby attractors---a car and its mirror image---are reminiscent of the two-level systems of disordered solids, where a system hops
  between two nearly degenerate configurations separated by a small barrier; the pairing here is probably a consequence of the horizontal-flip augmentation used in training.
  As in Stable Diffusion, the same attractors recur throughout our experiments from very different initial seeds and cycling amplitudes, and the candidates emerge from
  a single public checkpoint through the model's own dynamics, without the large-scale sample-and-search or the retraining of multiple models used by some prior CIFAR-10
  extraction attempts. The attractors in Fig.~\ref{fig:main_Figure6} and Supplementary Fig.~\ref{fig:Supplementary_Figure5} are a non-exhaustive selection: cycling exposes
  many more memorized attractors across seeds and amplitudes than we report here, both for CIFAR-10 and for Stable Diffusion (Figs.~\ref{fig:main_Figure4}
  and~\ref{fig:main_Figure5}). One caveat to keep in mind when comparing yield values across methods: training multiple models, each on a different subset of the
  data~\citep{carlini2023extractingtrainingdatadiffusion}, produces several independent landscapes, each with its own memorized minima. We therefore expect the yield of
  cyclic denoising to grow with the number of independently trained models it is applied to, so any direct comparison of recovery rates should control for the number of
  models.

  \textbf{Prompt-conditioned attractors and ultrastability after prompt removal.} Figure~\ref{fig:main_figure7} applies the same protocol with classifier-free guidance, using
  prompts previously associated with memorized Stable Diffusion images~\citep{carlini2023extractingtrainingdatadiffusion,webster2023reproducibleextractiontrainingimages}.
  Each such prompt rapidly drives the dynamics onto the corresponding known memorized image and holds it there, with different Gaussian seeds for the same prompt reaching the
  same attractor (Fig.~\ref{fig:main_figure7}(a,c)). Because these prompts target images \emph{independently} known to be memorized, their recovery as attractors is a
  positive control: it confirms directly that cyclic denoising can drive the dynamics onto genuinely memorized content. Removing the prompt and continuing to cycle
  unconditionally then probes their stability: for ``Ann Graham Lotz'', four of five trajectories remain locked in the same image for the entire $1000$-cycle unconditional
  continuation we ran---still absorbed when we stopped, rather than decorrelating---an ultrastable basin (Fig.~\ref{fig:main_figure7}(e,f)). A \emph{guessed} prompt---one not
  known to correspond to any training image---behaves differently. With ``Abbey Road album cover'' the dynamics still reach a basin that is stable under the prompt, but the
  decoded state varies slightly from cycle to cycle and never settles on the actual cover: it captures the concept rather than a stored image, and decorrelates within a few cycles
  once conditioning is removed (Fig.~\ref{fig:main_figure7}(g,h)). A second guessed prompt, ``Mona Lisa'', likewise settles under its prompt into a recognizable but
  non-verbatim, fluctuating concept attractor; not every prompt induces such ready absorption, but among several candidate prompts we tried, these two absorbed readily. 
  The two cases differ on two signatures: first, under the prompt, the known memories are sharp and verbatim, returning to essentially the same image from cycle to cycle,
  whereas the guessed prompts give softer, fluctuating renditions of the concept (Supplementary Fig.~\ref{fig:Supplementary_Figure6}); and second, after prompt removal,
  decorrelation time correlates positively with memorization, with the guessed concept basin decorrelating the fastest and the memorized images spanning a range of
  stabilities up to the ultrastable basins (Supplementary Fig.~\ref{fig:Supplementary_Figure7}).  Our unconditional extraction relies on this signal---long-lived, ultrastable absorption---to flag memorized candidates without any prompt; that the deepest known memories
  stay ultrastable without the prompt is a ground-truth check that the signal is sound.

\section{Discussion}
Cyclic denoising changes how we should think about memorization in diffusion models. Prior extraction attacks largely treat the model as a generator of independent samples and then search for memorized examples by prompting, clustering, or comparing against the training set. Our results show that memorized images can instead appear as dynamical attractors: not merely rare outputs, but stable states that the model can repeatedly regenerate under cyclic perturbation. Standard sampling may therefore substantially underestimate memorization risk, because some training examples are hidden in deep basins that are rarely visited unless the model is driven.

More broadly, our results suggest that memorization should be studied as a stability property of the learned generative dynamics.  In prompt-conditioned settings, stability after prompt removal already provides a targeted test: images that remain locked without conditioning are more strongly memorized
than concept basins stabilized only by the prompt. Training examples that survive strong cyclic perturbations behave like deep basins in the model landscape. Preventing memorization may therefore require more than reducing the probability of reproducing a training image under ordinary sampling; it may require eliminating or weakening the ultrastable attractors that store those images.

\section{Limitations}
  Our attack requires sampler-level control over the noising and denoising steps, and therefore does not apply directly to standard text-to-image API endpoints that expose
  only final samples. Coverage requires a more quantitative investigation: while some basins clearly capture more trajectories than others in our runs, we have not systematically characterized the shape of the attractor distribution or the fraction of the memorized set that any single sweep recovers. The recovered attractors depend on choices such as the cycling amplitude, scheduler, number of cycles, and diversity within initialization ensembles; systematic
  optimization of these choices is an important direction for future work.

\section{Broader Impacts}
As an auditing tool, cyclic denoising lets practitioners and third parties identify training images that a deployed diffusion model has memorized, including copyrighted photographs, watermarked content, brand logos, and privacy-sensitive material scraped from the open web. The recovered attractor set provides a model-specific fingerprint of repeatedly encoded images that reveals information about the data distribution and the training pipeline. Because the protocol uses no captions, training-set access, or weight inspection, it can be applied as a dataset-agnostic audit of open-weight diffusion models that would otherwise be hard to inspect.

The same protocol is also an extraction attack. An adversary with sampler-level access could recover sensitive training content without the prompts, captions, or training-set lookups that bound caption-based pipelines, and the attack does not depend on knowing what to look for: any sufficiently deep basin will surface its memorized content under sustained cyclic forcing. Memorization rates measured under ordinary sampling may therefore understate the leakage that a determined adversary can elicit through cyclic perturbation, and absorbing-state structure should be treated as part of the threat model for diffusion-model deployment.

\section*{Acknowledgments}
This work was supported by a grant from the Simons Foundation [MPS-T-MPS-00839534, MET] (RS, SM). 
We gratefully acknowledge the use of computational resources and consultation support provided by NYU IT High Performance Computing. We also thank Prof. David J. Heeger for many fruitful discussions.

\clearpage

\bibliographystyle{plainnat}
\bibliography{references}


\appendix
\section{Supplementary material}
\begin{figure}[htpb]
    \centering
    \includegraphics[width=1.0\linewidth]{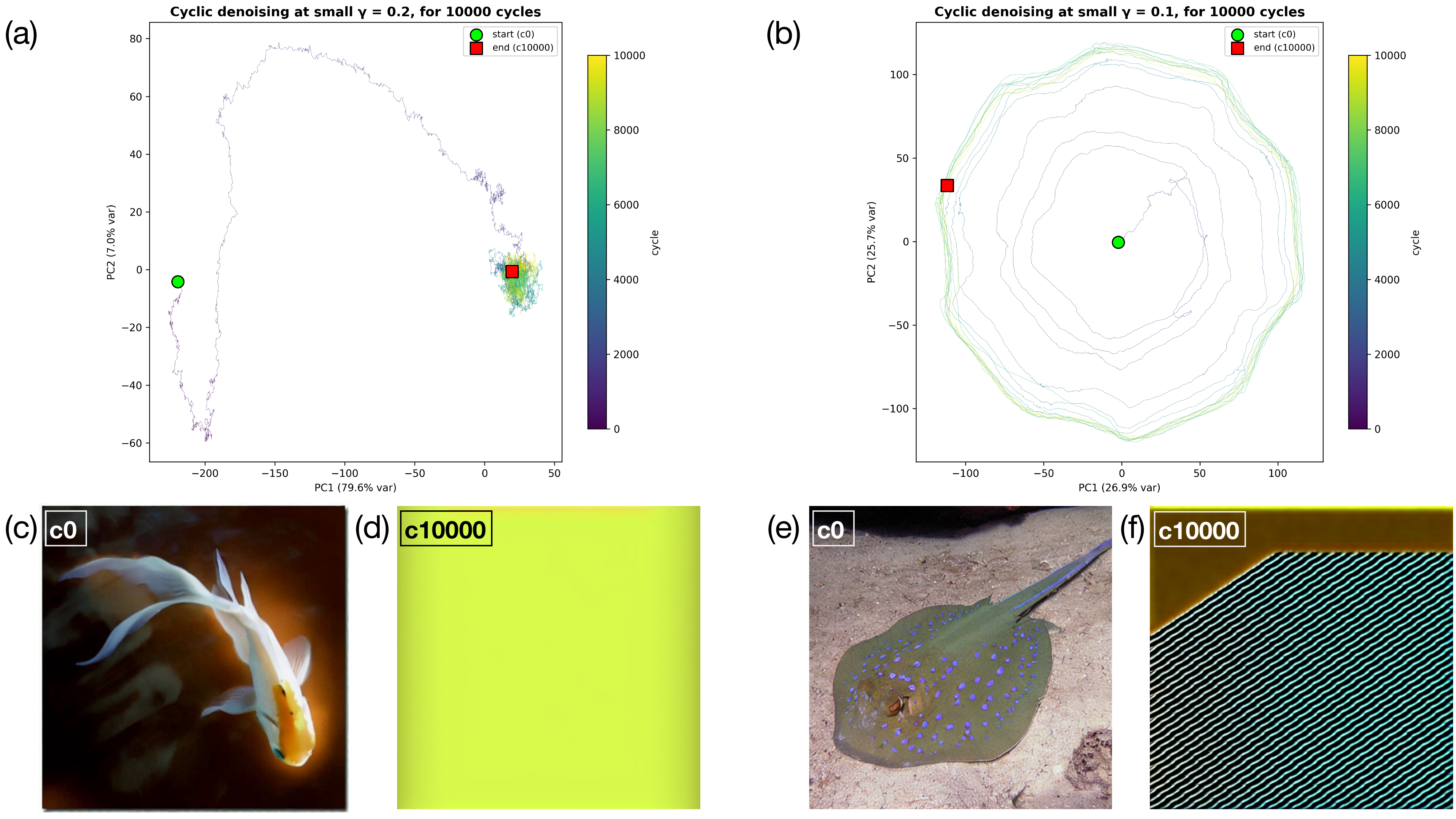}
    \caption{\textbf{Two kinds of attractor reached by low-amplitude unconditional cycling in Stable Diffusion v1.4: trivial fixed points and limit cycles.}
    (a,b) Two-dimensional PCA projections of stroboscopic cyclic-denoising trajectories, each run for $10^4$ cycles; line color encodes cycle number, the green circle marks
  the start ($c_0$) and the red square the end ($c_{10000}$), and the PCA basis is fit on each trajectory individually.
    (a) At $\gamma=0.2$ the trajectory drifts away from its initial latent and collapses into a small region of the PCA plane---a trivial absorbing \emph{fixed point} (up to
  small stochastic jitter), where the decoded image stays near-monochromatic and stops rearranging from cycle to cycle.
    (b) At $\gamma=0.1$ the trajectory instead settles onto a closed orbit, traversing nested, near-concentric loops in the PCA plane: a \emph{limit cycle} in which the state
  returns periodically rather than freezing.
    (c--f) Decoded latents at the start and end of each trajectory. (c) Start and (d) end of (a): an initial natural image relaxes to a featureless, near-monochromatic frame.
  (e) Start and (f) end of (b): the image converges to a periodic, Turing-like stripe pattern. The endpoint in (f) is not static: the (spatially periodic) stripe pattern
  travels across the frame as cycling proceeds, and it is this periodicity in cycle number---the wave returning to itself every fixed number of cycles---rather than the
  spatial periodicity of any single frame, that is the real-space signature of the closed orbit in (b). This is reminiscent of the classic observation that repeatedly blurring and then sharpening an image produces Turing-like patterns through an effective reaction--diffusion
  dynamics~\citep{bridges2015Turing}; in cyclic denoising the smoothing and sharpening are instead carried out by the learned diffusion kernel.
    Side-by-side animations of both trajectories and their decoded latents are provided in Supplementary Movies~2 and~3.
    Limit cycles are considerably rarer than decay to trivial fixed points. In both regimes the low-amplitude dynamics explore little of the landscape, in contrast to the
  more space-filling transients seen at larger amplitudes.}
    \label{fig:supplementary_Figure2}
  \end{figure}

  All supplementary videos are available at \url{https://rishabh-tifr.github.io/cyclic-denoising/movies}. Each animates a stroboscopic cyclic-denoising trajectory in Stable
  Diffusion v1.4, showing the 2D PCA projection of the latent (colored by cycle number) alongside the simultaneously decoded image:
  \begin{itemize}[leftmargin=*,itemsep=2pt,topsep=2pt]
    \item \textbf{Movie~1} --- basin hopping at $\gamma=0.86$, the trajectory of Figs.~\ref{fig:main_Figure1} and~\ref{fig:main_Figure2}.
    \item \textbf{Movie~2} --- collapse into a trivial absorbing fixed point at $\gamma=0.2$ (Supplementary Fig.~\ref{fig:supplementary_Figure2}(a)).
    \item \textbf{Movie~3} --- a low-amplitude limit cycle at $\gamma=0.1$ (Supplementary Fig.~\ref{fig:supplementary_Figure2}(b)).
    \item \textbf{Movie~4} --- a second, less regular limit cycle at $\gamma=0.1$ (additional example, not shown in the paper).
  \end{itemize}

\begin{figure}[htpb]
  \centering
  \includegraphics[width=1.0\linewidth]{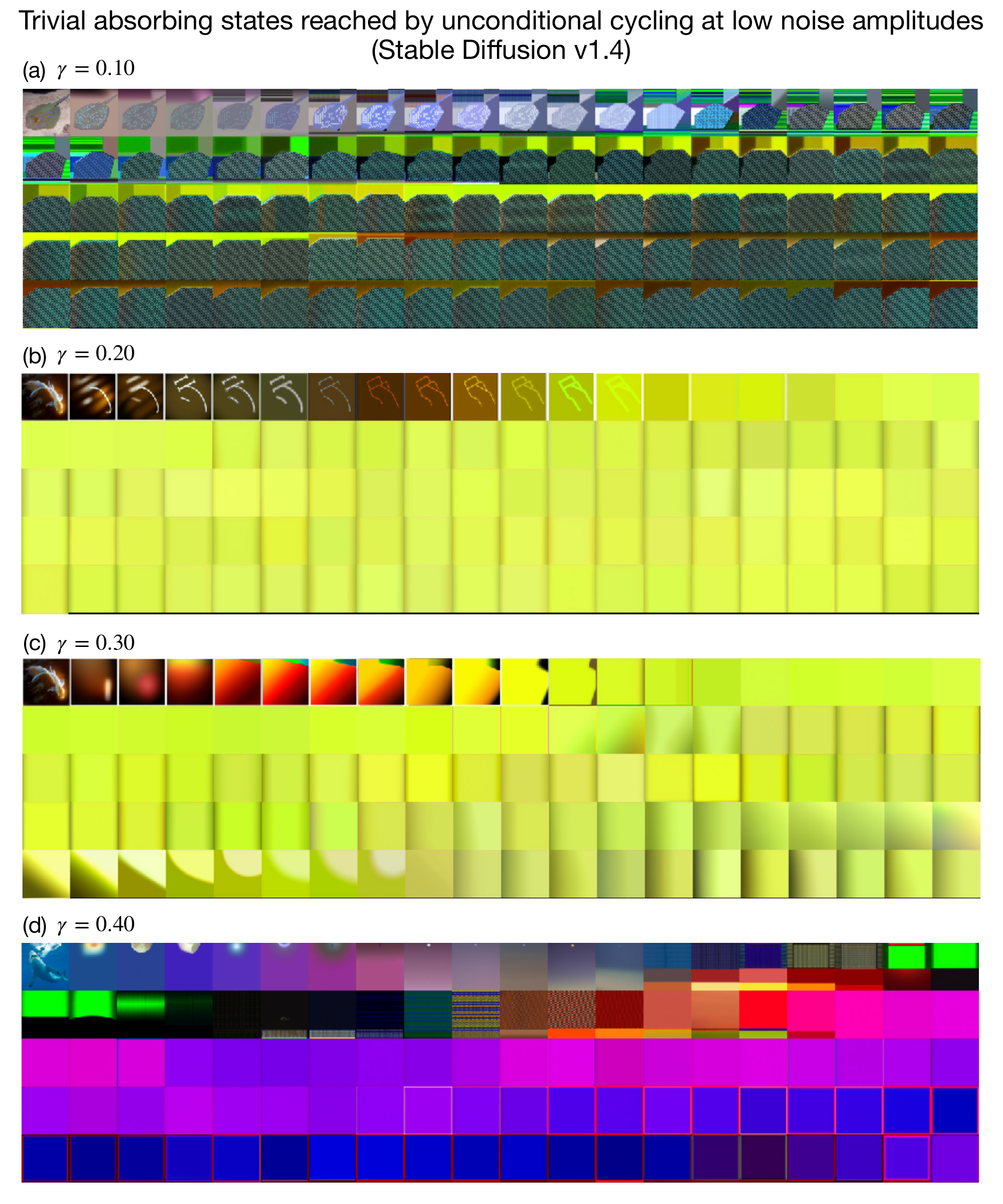}
   \caption{\textbf{Trivial absorbing states from low-amplitude unconditional cycling in Stable Diffusion v1.4.}
Unconditional cyclic denoising initialized from ImageNet test images at low noise amplitudes: (a) $\gamma=0.1$, (b) $\gamma=0.2$, (c) $\gamma=0.3$, and (d) $\gamma=0.4$. Each trajectory is run for $10{,}000$ cycles, with every 100th cycle shown. At these amplitudes, the perturbation is too weak to induce large-scale rearrangements, and the dynamics collapse into trivial absorbing states such as saturated patterns, monotonic fills, or simple geometric structures.}
  \label{fig:supplementary_Figure1}
\end{figure}

\begin{figure}[htpb]
      \centering
      \includegraphics[width=1.0\linewidth]{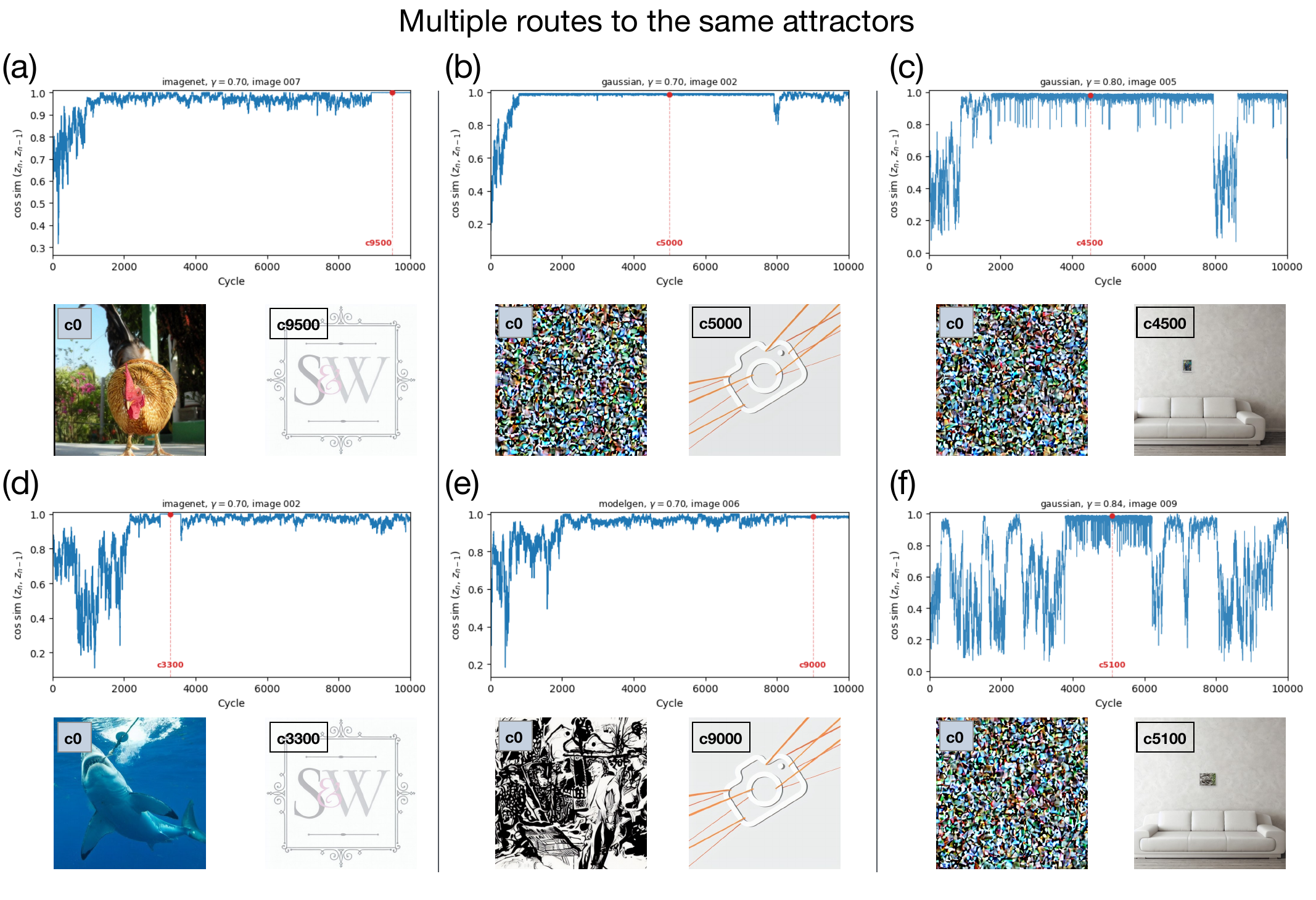}
      \caption{\textbf{Multiple routes to the same attractors.}
      Cyclic denoising recovering the same memorized states as attractors across different initial conditions and noising amplitudes $\gamma$. In each panel, the top trace is the consecutive stroboscopic cosine similarity, $\cos(z_n, z_{n-1})$, along a 10{,}000-cycle trajectory; values near 1 mark an \emph{absorbing region} in which successive cycles return to essentially the same latent. The red marker indicates a cycle deep within such a region, and the two images below decode the latent at the initial cycle (c0, left) and at the marked cycle (right). Within each column, the two rows begin from different conditions yet are captured by the \emph{same} attractor, isolating one axis of variation at a time: \textbf{(a, d)} two distinct ImageNet seeds at $\gamma = 0.70$; \textbf{(b, e)} two distinct seed ensembles---Gaussian and model-generated---at $\gamma = 0.70$; \textbf{(c, f)} two distinct amplitudes, $\gamma = 0.80$ and $\gamma = 0.84$, both from Gaussian seeds. ``ImageNet'', ``model-gen'', and ``Gaussian'' denote how the initial seeds were produced---VAE-encoded ImageNet validation images, samples generated by the model, and pure Gaussian latents, respectively---and ``image NNN'' is the index of the trajectory within its ensemble. The same memorized images recur across these independent runs, indicating that the cyclic dynamics explore a single underlying landscape: the seeds act only as starting anchors, while the amplitude $\gamma$ sets the effective step size of the exploration. A given memorized image can therefore be reached from a range of seeds and amplitudes; for clarity we show only two representative routes to each of the three attractors.
      }
      \label{fig:Supplementary_Figure3}
  \end{figure}

  \begin{figure}[htpb]
      \centering
      \includegraphics[width=1.0\linewidth]{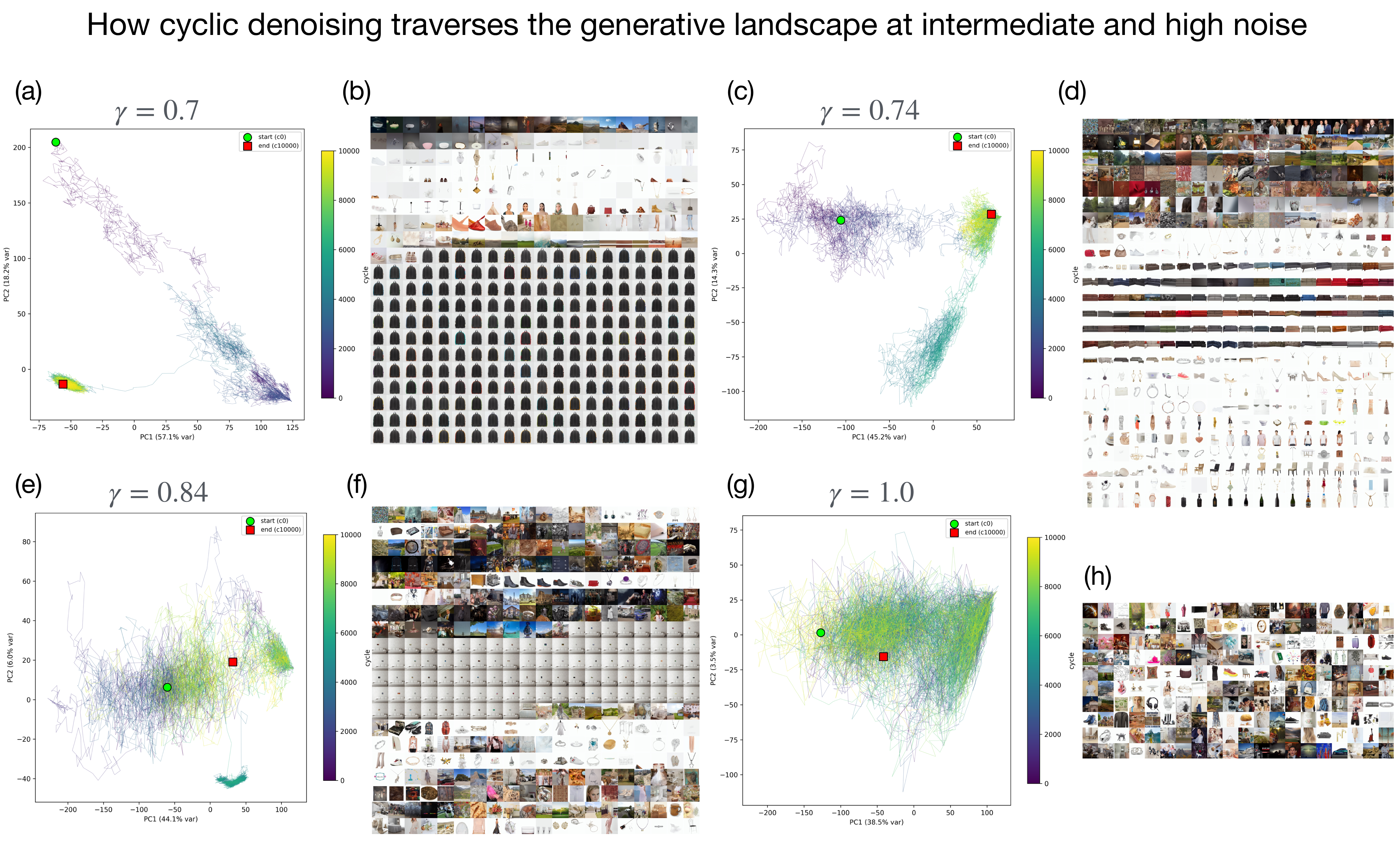}
      \caption{\textbf{How cyclic denoising traverses the generative landscape at intermediate and high noise amplitudes.}
      For each amplitude $\gamma$ we show a per-trajectory 2D PCA projection of a 10{,}000-cycle latent trajectory (color encodes cycle number; the green circle marks the
  initial latent c0 and the red square the final cycle c10000) together with a montage of decoded latents sampled along the same trajectory in cycle order.
      \textbf{(a)}~$\gamma = 0.7$: in the PCA projection the trajectory collapses into a single deep attractor and dwells there for the remainder of the run.
      \textbf{(b)} the corresponding decoded montage (every 25th cycle) converges onto, and then repeatedly reproduces, the same memorized bag.
      \textbf{(c)}~$\gamma = 0.74$: no single deep attractor forms, yet the projection resolves into a few diffuse but separated clusters, the trajectory dwelling within one
  before hopping to the next.
      \textbf{(d)} the corresponding decoded montage (every 20th cycle) shows these clusters to be \emph{semantically coherent}: an early, possibly transient region
  containing a mix of scenes and human portraits/photographs, followed by a region of sofa images---which change from cycle to cycle but all remain sofas, and therefore
  collapse into a single PCA cluster---and finally, a region of catalog-style product images. Thus, even in the absence of an absorbing state, cyclic denoising traverses the landscape
  in a semantically segmented fashion, dwelling in shallow basins that each encode a coherent concept and hopping between them. This semantic structuring of the dynamics is
  itself intriguing: it holds generically---even on runs that never settle onto a deep memorized attractor---and points to cyclic denoising as a route to new methods
  for probing the semantic organization of a model's landscape.
      \textbf{(e)}~$\gamma = 0.84$: the PCA projection shows the trajectory again captured by a deep basin partway through the run before escaping it.
      \textbf{(f)} the corresponding decoded montage (every 25th cycle) identifies this basin as the white-sofa scene.
      \textbf{(g)}~$\gamma = 1.0$: at the (almost) full-noise limit the projection shows no distinct clusters.
     \textbf{(h)} the decoded latents (every 50th cycle) are essentially independent from cycle to cycle, since each cycle re-noises the latent almost completely---though not
  entirely: Stable Diffusion's noise schedule has a non-zero terminal SNR, so even at $\gamma=1$ a small fraction of the previous latent is carried over rather than a true
  reset to pure noise.
      Across amplitudes, intermediate $\gamma$ thus exposes a hierarchy of basins---deep absorbing attractors that correspond to memorized images, and shallow semantic basins
  between which the dynamics wander---whereas $\gamma \to 1$ essentially erases this structure.
      }
      \label{fig:Supplementary_Figure4}
  \end{figure}
  
\begin{figure}[htpb]
  \centering
  \includegraphics[width=1.0\linewidth]{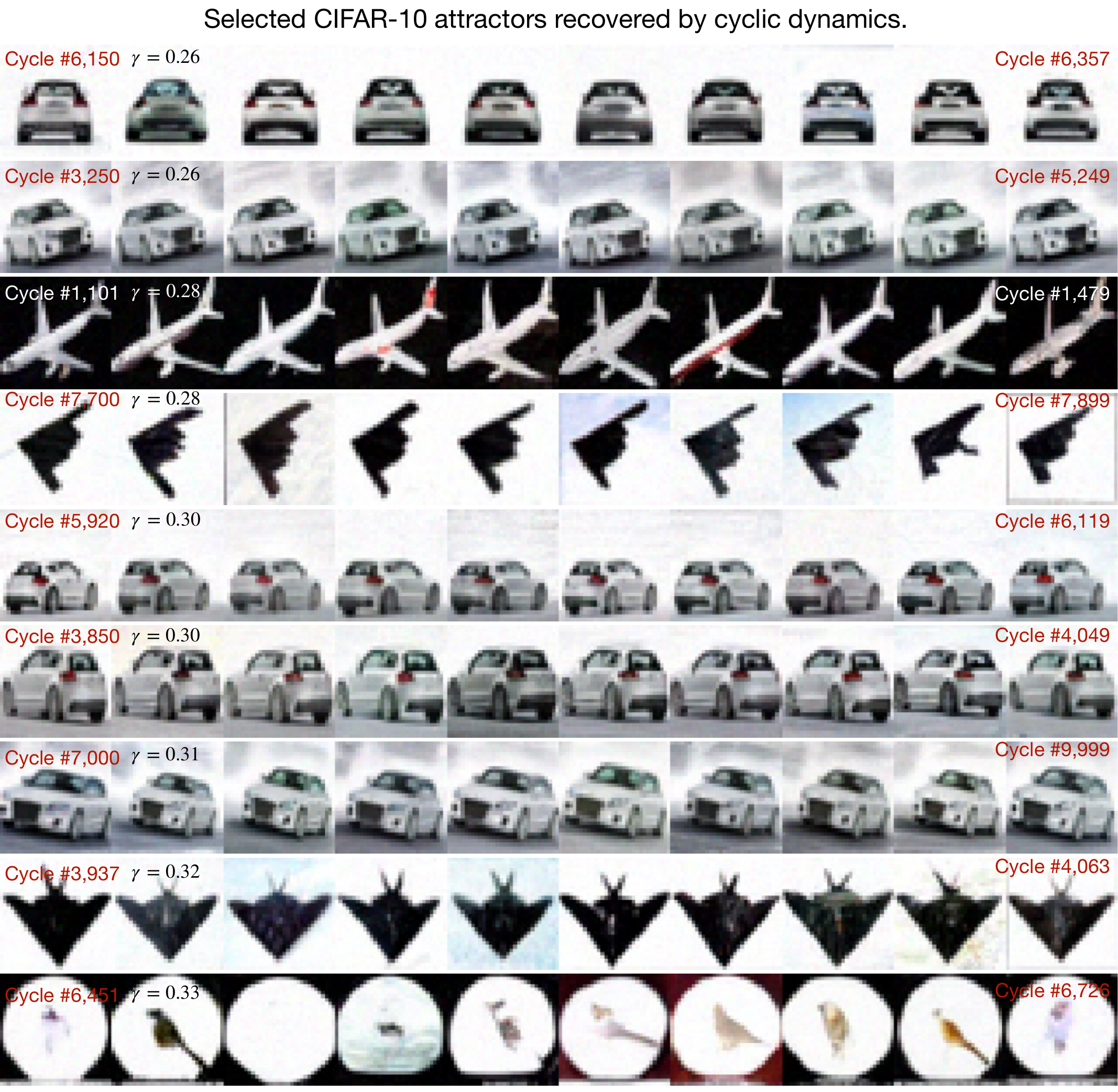}
  \caption{\textbf{Selected CIFAR-10 attractors recovered by unconditional cyclic dynamics.}
Each row shows ten equispaced stroboscopic snapshots from a single $10{,}000$-cycle trajectory during the residence time of one basin, with the cycle range and $\gamma$ indicated. Several attractors match CIFAR-10 training examples previously identified as memorized by \citet{carlini2023extractingtrainingdatadiffusion} in independently trained diffusion models, suggesting that these are memorization-prone images that recur across training runs. Cyclic denoising recovers these images directly from a publicly available CIFAR-10 DDPM checkpoint, without prior knowledge of (or access to) the training set or any post-hoc training-set search.}
  \label{fig:Supplementary_Figure5}
\end{figure}

\begin{figure}[htpb]
  \centering
  \includegraphics[width=1.0\linewidth]{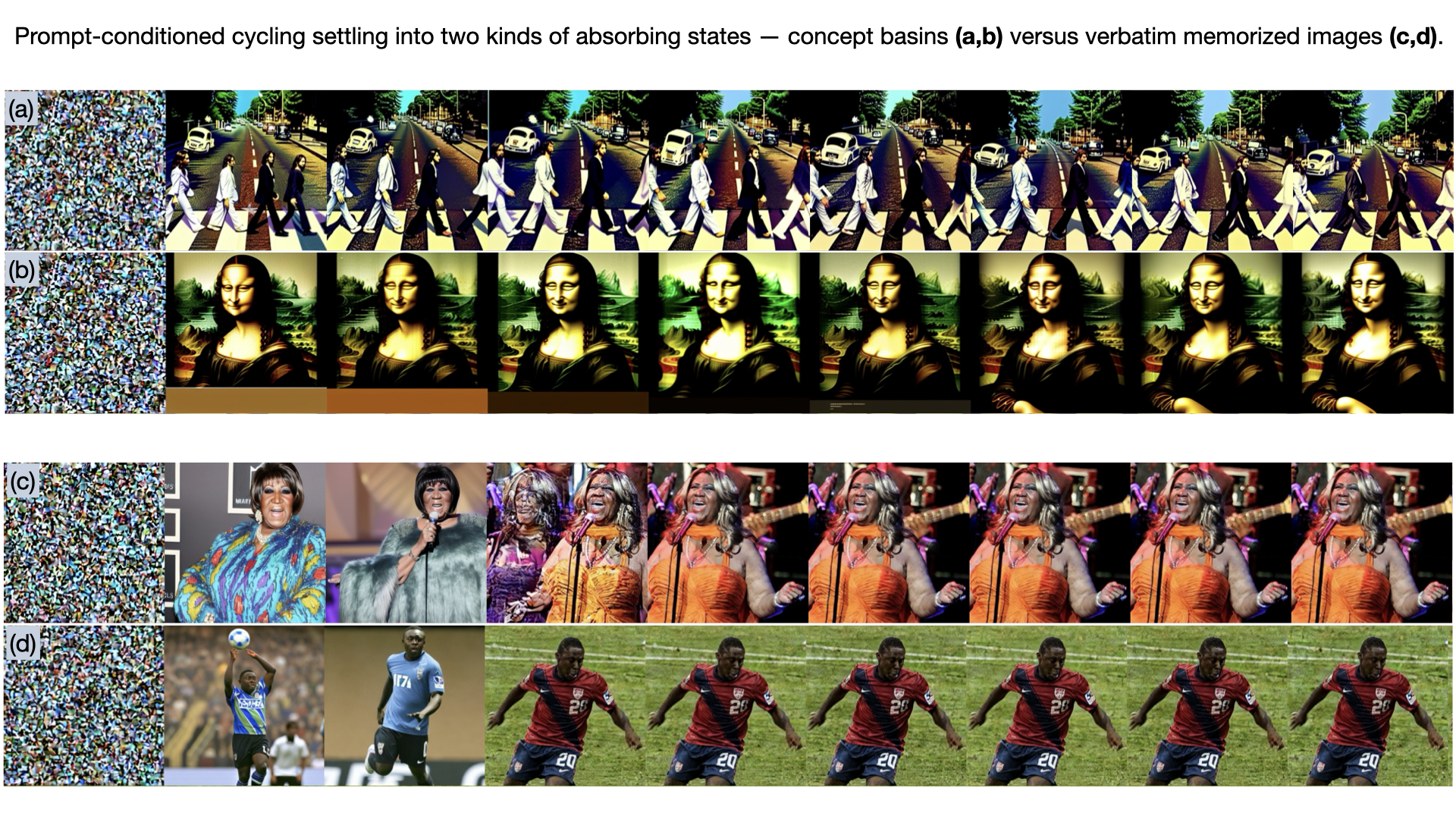}
   \caption{\textbf{Prompt-conditioned cycling settling into two kinds of absorbing
    states: concept basins (a,b) versus verbatim memorized images (c,d).} Each row shows
    the decoded latent across cycles, from a noise seed (left) to the absorbing state.
    Concept basins---\textbf{(a)} ``Abbey Road album cover'' and \textbf{(b)} ``Mona
    Lisa''---settle into a recognizable but \emph{crude} rendering: conceptually correct,
    yet not photorealistic to any single training image. Memorized prompts---\textbf{(c)}
    Aretha Franklin and \textbf{(d)} Freddy Adu---settle into a \emph{sharp, verbatim}
    copy of one specific training image.}
  \label{fig:Supplementary_Figure6}
\end{figure}

\begin{figure}[htpb]
    \centering
    \includegraphics[width=1.0\linewidth]{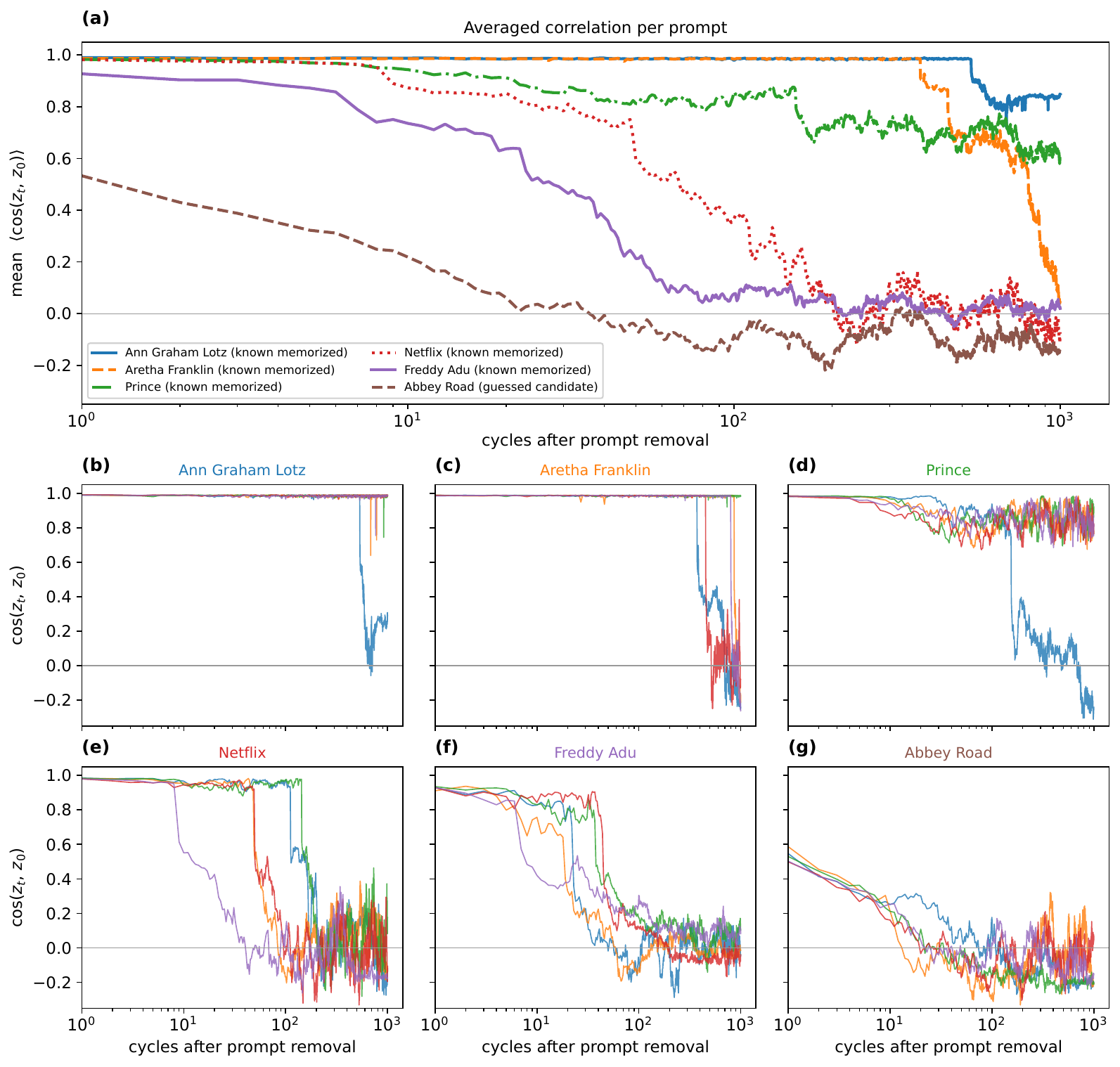}
    \caption{\textbf{Decorrelation of prompt-conditioned absorbing states after prompt
    removal ($\gamma = 0.78$).} Each prompt is first cycled to a prompt-conditioned
    absorbing state; the prompt is then removed and unconditional cyclic denoising
    continues for 1000 cycles at fixed amplitude $\gamma = 0.78$, while we track the
    correlation $\cos(z_t, z_0)$ between each cycle and the absorbed image at the moment
    of removal (logarithmic cycle axis; 5 seeds per prompt). \textbf{(a)} Seed-averaged
    correlation for all six prompts. \textbf{(b--g)} The corresponding per-seed ensembles:
    \textbf{(b)} ``Ann Graham Lotz'' ($4/5$ seeds survive $10^3$ cycles);
    \textbf{(c)} ``Aretha Franklin Files \$10 Million Suit Over Patti LaBelle Fight Story
    On Satire Website'' ($1/5$);
    \textbf{(d)} ``Prince Reunites With Warner Brothers, Plans New Album'' ($4/5$);
    \textbf{(e)} ``Netflix Hits 50 Million Subscribers'' ($0/5$);
    \textbf{(f)} ``Freddy Adu Signs For Yet Another Club You Probably Don't Know'' ($0/5$);
     \textbf{(g)} ``Abbey Road album cover'' ($0/5$), a guessed prompt that forms a
    prompt-stabilized concept basin rather than a confirmed memorized image. The concept
    basin (g) has the shortest decorrelation time, decaying fastest once the prompt is
    removed, while the known memorized images (b--f) show a range of stabilities at this
    amplitude.}
    \label{fig:Supplementary_Figure7}
  \end{figure}

\begin{table}[h]
\caption{\textbf{Reverse-image-search matches for recovered Stable Diffusion attractors.}
URLs correspond to representative web matches used for post-hoc attribution and verification. Accessed May 2026.}
\label{tab:Supplementary_Table1}
\centering
\footnotesize
\begin{tabular}{p{0.30\linewidth}p{0.62\linewidth}}
\toprule
 Match & Source URL \\
\midrule
Yellow chair and white sofa scene & \url{https://fineartamerica.com/featured/1-president-john-f-kennedy-everett.html?product=poster} \\
Nalli  & \url{https://www.justdial.com/jdmart/Barmer/Bangalore-Silk-Fabric/pid-2011643866/9999P2982-2982-190706194132-H2Z4} \\
SET  & \url{https://www.fashionsauce.com/brands/set-by-ouiset-stockists} \\
PhotoShelter unavailable placeholder & \url{https://ssl.c.photoshelter.com/img/photoshelter_unavailable.jpg} \\
Tadashi Shoji  & \url{https://www.tadashishoji.com/3k896mx-black-embroidered-lace-v-neck-dress} \\
Sweetpea and Willow (SW logo) & \url{https://www.homify.co.uk/professionals/2824148/sweetpea-and-willow-london-ltd} \\
Camera Placeholder & \url{https://image.ceneostatic.pl/data/products/113506095/p-the-north-face-kurtka-damska-apex-flex-shell-gtx-trellis-green.jpg} \\
Ogio Alpha Convoy 525 backpack & \url{https://procare.gr/6979-large_default/ogio-alpha-convoy-525-backpack-charcoal.jpg} \\
Ogio Shadow Fuse stand (Blue golf bag)  & \url{https://static.golfballs.com/C/800x800/Products/Legacy/47/Ogio-Shadow-Fuse-304-Stand-Bag_RYL_raw.webp} \\
Tote bag & \url{https://fineartamerica.com/featured/midwestern-cotton-candy-l-p.html?product=tote-bag} \\
Ogio Convoy 320 backpack (Black bag) & \url{https://images.squarespace-cdn.com/content/v1/583335f52994ca7d6adc6f3d/1547569522591-MWE7TCPD2DZN4CVYWA5T/Ogio-convoy-320-backpack-review-01.jpg?format=1500w} \\
\bottomrule
\end{tabular}
\end{table}


\newpage
\clearpage

\end{document}